\documentclass[conference]{IEEEtran}
\IEEEoverridecommandlockouts

\usepackage{cite}
\usepackage{adjustbox}
\usepackage{amsmath,amssymb,amsfonts}
\usepackage{graphicx}
\usepackage{booktabs}
\usepackage{textcomp}
\usepackage{xcolor}
\usepackage{tabularray}
\usepackage{algorithm}
\usepackage{gensymb}
\usepackage{algpseudocode}
\usepackage{comment}
\usepackage{threeparttable}

\usepackage{caption}
\usepackage{subcaption}
    
\begin{document}

\title{Unsupervised Semantic Segmentation in Synchrotron Computed Tomography with Self-Correcting Pseudo Labels}

\author{
\IEEEauthorblockN{Austin Yunker$^1$, Peter Kenesei$^2$, Hemant Sharma$^2$, Jun-Sang Park$^2$, Antonino Miceli$^2$, Rajkumar Kettimuthu$^1$}
$^1$Data Science and Learning Division, Argonne National Laboratory, Lemont, IL USA\\
$^2$X-Ray Science Division, Argonne National Laboratory,
Lemont, IL USA \\
\{ayunker, kenesei, hsharma, parkjs, amiceli, kettimut\}@anl.gov
}

\maketitle

\begin{abstract}
X-ray computed tomography (CT) is a widely used imaging technique that provides detailed examinations into the internal structure of an object with synchrotron CT (SR-CT) enabling improved data quality by using higher energy, monochromatic X-rays. While SR-CT allows for improved resolution, time-resolved experimentation, and reduced imaging artifacts, it also produces significantly larger datasets than conventional CT. Accurate and efficient evaluation of these datasets is a critical component of these workflows; yet is often done manually representing a major bottleneck in the analysis phase. While deep learning has emerged as a powerful tool capable of providing a wide range of purely data-driven solutions, it requires a substantial amount of labeled data for training and manual annotation of SR-CT datasets is impractical in practice. In this paper, we introduce a novel framework that enables automatic segmentation of large, high-resolution SR-CT datasets by eliminating the need to hand label images for deep learning training. First, we generate pseudo labels by clustering on the voxel values identifying regions in the volume with similar attenuation coefficients producing an initial semantic map. Afterwards, we train a segmentation model on the pseudo labels before utilizing the Unbiased Teacher approach to self-correct them ensuring accurate final segmentations. We find our approach improves pixel-wise accuracy and mIoU by 13.31\% and 15.94\%, respectively, over the baseline pseudo labels when using a magnesium crystal SR-CT sample. Additionally, we extensively evaluate the different components of our workflow including segmentation model, loss function, pseudo labeling strategy, and input type. Finally, we evaluate our approach on to two additional samples highlighting our framework’s ability to produce segmentations that are considerably better than the original pseudo labels.

\end{abstract}    
\section{Introduction}
\label{sec:intro}

X-ray computed tomography (CT) is a popular imaging technique that provides detailed examinations into the internal structure of an object \cite{buzug2008computed}. Due to its non-invasive and -destructive properties, CT has found widespread uses in various fields including materials science \cite{omori2023recent}, biological and medical research \cite{hiriyannaiah1997x}, earth and environment sciences \cite{taina2008application}, and structural health monitoring \cite{nateghi2024health}. By utilizing extremely bright, highly coherent monochromatic X-rays, synchrotron CT (SR-CT) allows for sub-micron resolution, time-resolved experimentation, and reduced imaging artifacts \cite{THOMPSON1984319, KIM198744, KASTNER2010599}.  Applications of SR-CT include a diverse range of scientific disciplines including additive manufacturing \cite{PEGUES2021159505}, disease imaging \cite{tavakoli2020comparison}, and drug delivery \cite{patterson2020novel}. While conventional CT is ubiquitous in practice, access to SR-CT is often restricted due to its high cost and requires proposal-based access to a beamline at facilities.  

Although SR-CT achieves improved data quality at higher resolutions, it also produces significantly larger datasets than conventional CT with the work in \cite{hidayetouglu2020petascale} generating a 4.3 terabyte mouse brain volume with 9K$\times$11K$\times$11K voxels. One of the factors that directly affects volume size is detector resolution \cite{hampai2025x}. Given that pixel sizes range from 0.1-1$\mu$m, covering a millimeter-size area requires detectors to record 2k-4k pixels per dimension resulting in $2048\times2048$ to $4096\times4096$ image size. As an example, the work in \cite{yakovlev2022wide} measured a field of view of 5.0 mm $\times$ 3.5 mm using 0.5$\mu$m voxel size resulting in a volume of $7096 \times 10000\times 10000$ voxels. In addition to detector resolution, sample size plays an important role in the size of the generated data. If the sample diameter fits into the beam, but is long vertically (axially), several consecutive layers of the sample are scanned with the resulting volumes stitched together. Furthermore, if the sample diameter is larger than the beam, several horizontally shifted scans are needed to cover the whole sample diameter, which are then stitched together before reconstruction. By having N consecutive layers and M horizontally shifted scans, samples that require both approaches yield a final volume N$\times$M$^2$ times larger than a single scan \cite{khounsary2013high}.

After generation, accurate and efficient evaluation of the resulting volumes plays a critical role in these workflows as their results are used to asses organ changes and diagnose diseases in the medical domain \cite{han2017volume}, quantify bulk materials necessary for optimizing material properties in materials science \cite{omori2023recent}, and identify pore networks and root architectures in environmental science \cite{ziegler2021computed}. Part of the evaluation stage includes segmenting each pixel in the CT scan effectively identifying different tissues, organs, or structures. Done manually, this segmentation approach requires a domain expert representing a significant bottleneck in the workflow with the work in \cite{choi2024development} reporting an average of 111 minutes to manually segment the psoas muscle in a CT scan. Recently, deep learning has emerged as a powerful tool capable of providing a wide range of purely data-driven solutions with applications ranging from cancer diagnosis \cite{munir2019cancer} to material characterization \cite{nateghi2024health}. Applied to segmentation, deep learning offers a significant reduction in evaluation time for segmentation tasks with performance at or above human level \cite{mao2023cross}. While segmentation methods have been successfully applied to the field of CT, most notably medical CT \cite{isensee2021nnu, ma2024segment}, these models contain millions of parameters that require a large amount of high-quality annotated data during training. For SR-CT samples, commonly generated at more than a terabyte \cite{nikitin2020dynamic}, manual labeling is unrealistic in practice and requires an alternative approach.

To overcome this issue, researchers have explored semi-supervised methods that focus on training a model using a small amount of labeled data while utilizing a massive dataset of unlabeled images to improve performance \cite{van2020survey, chen2021semi, wang2022semi}. Self-training via pseudo labeling \cite{lee2013pseudo, xu2022semi, yang2022st++} has developed into a mainstream technique within this field. This method trains a model in a fully-supervised fashion on the labeled images and then employs the model to generate pseudo labels for all unlabeled images. Afterwards, the model is re-trained from scratch using all ground-truth and pseudo labels. However, it was observed in \cite{arazo2020pseudo} that using pseudo labels will trigger the confirmation bias resulting in the model overfitting to the incorrect labels. While semi-supervised methods provide a promising alternative to manual curation, additional challenges arise when applying them to SR-CT datasets. During the experimental setup, various factors relating to the x-ray source, environmental conditions, sample properties, and data processing algorithms can introduce noise and artifacts into the images further raising concern on the quality of the generated labels \cite{buzug2008computed}. Additionally, unlike medical CT, which involves repeated scanning of similar samples, SR-CT is applied to a wide variety of specimens across scientific disciplines collected under different experimental settings. This severely limits the utility of pre-trained models and/or previously labeled datasets as models often exhibit poor generalizability to unseen samples \cite{wang2022generalizing}. 

In this paper, motivated by the challenges of segmenting large, high-resolution SR-CT datasets, we propose a novel three-stage framework that enables automatic segmentation and completely eliminates the need for collecting manual labels. First, we generate initial pseudo labels to learn from. By clustering images based on voxel values, we can identify semantically coherent structures that likely belong to the same class. In the second stage, we train a segmentation model using these labels allowing the model to learn simple characteristics of the data regarding structures with similar attenuation coefficients. Finally, to correct for any noise in the pseudo labels and instill a more holistic understanding of the data into the model, we adapt the Unbiased Teacher approach \cite{liu2021unbiased} for semantic segmentation. We remark that, throughout the three stages, we rely solely on the raw images requiring no manual labels allowing for a fully unsupervised framework. Quantitative results on a real-world magnesium crystal SR-CT dataset collected at APS demonstrate that the proposed approach improves the pixel-wise accuracy and mean Intersection over Union (mIoU) by 13.31\% and 15.94\%, respectively, over the initial pseudo labels. Additionally, we rigorously evaluate the different components of our framework including segmentation model, loss function, pseudo label strategy, and input dimension before making use of class activation maps \cite{selvaraju2017grad} to explore model interpretability. Finally, we extend our approach to two additional real-world samples highlighting our framework's ability to produce segmentations that are considerable better than the original pseudo labels. To summarize, our key contributions include:
\begin{itemize}
    \item We propose a novel framework to address the challenge of segmenting large, high-resolution SR-CT datasets that overcomes the need for manual labels. 
    \item Our framework operates in three stages: a pseudo label generation stage, an initial learning stage, and a self-correction stage.
    \item To correct for any noise and imaging artifacts in the initial pseudo labels, we adapt the Unbiased Teacher approach for unsupervised semantic segmentation.
    \item By utilizing class activation maps, we show that our final model contains a more holistic understanding of the data beyond that learned in the initial training phase.  
\end{itemize}

The remainder of this paper is organized as follows. Section \ref{sec:rw} provides the related work on segmentation methods for CT including pseudo labeling and vision foundation models. In Section \ref{sec:method}, we introduce the necessary background on CT and present our framework for unsupervised segmentation of SR-CT. We perform the experimental results in Section \ref{sec:exp} and discuss their implications in Section \ref{sec:discussion}. Finally, we conclude in Section \ref{sec:conc}.

\section{Related Work}
\label{sec:rw}

\subsection{Semantic segmentation for CT.} 

Various types of CT have evolved over the years ranging from micro-, nano-, ultrasound, cone-beam, synchrotron, and helical each with specific characteristics that have benefited a wide range of scientific disciplines \cite{buzug2008computed}. Regardless of type, CT datasets are regularly segmented for identification, localization, and characterization with deep learning-based approaches offering a tremendous reduction in analysis time. The researchers in \cite{cooley2021semantic} demonstrated that a convolutional neural network (CNN) was capable of accurately segmenting the mineralized tissues of mice jaws using micro-CT reconstructions. Badran et al. utilized a synchrotron micro-tomography beamline to analyze matrix cracks generated during in situ tensile loading of fiber-reinforced ceramic composites. They were able to successfully train FCDenseNet to automate the segmentation of the cracks as well as distinguish different phases in the material \cite{badran2020automated}. The researchers in \cite{tsamos2023synthetic} introduced a method for generating synthetic data for segmenting six-phase Al-Si alloy composite reinforced with ceramic fibers and particles requiring carefully crafted data augmentations and a specifically designed UNet (Triple UNet) with a multi-view forwarding strategy to achieve accurate segmentations. Recently, Manchester et al. demonstrated in \cite{manchester2025leveraging} the potential of transforming ex situ data for model training achieving segmentation performance that matched human reliability enabling the analysis of time-resolved data generated during in situ experiments. Although the above approaches have shown deep learning can achieve human level performance for segmenting CT datasets in a fraction of the time, their need to first curate a dataset of high-quality ground truth labels for training significantly limits their widespread use as these models often suffer from low generalizability \cite{schwonberg2025domain}. This work differs from those above by developing a workflow that utilizes automatically generated pseudo labels allowing it to be used for any SR-CT dataset.

\subsection{Pseudo labeling for semantic segmentation.} To address that lack of high-quality ground truth labels for training deep learning models, researchers have turned to semi-supervised learning to harness the potential of unlabeled data. Within this paradigm, pseudo labeling has emerged as a straightforward and widely applicable approach that aims to develop a model that can effectively utilize limited labeled data while extracting valuable insights from a vast amount of unlabeled data. One of the earliest works proposed assigning labels based on the class which has the maximum predicted probability \cite{lee2013pseudo}. Recent methods are generally split into model structure \cite{tarvainen2017mean, xu2022semi}, pseudo label refinement \cite{sohn2020fixmatch, li2021residual}, data enhancement \cite{yuan2021simple, chen2021complexmix}, and optimization improvement \cite{wang2022learning, kong2023pruning}. Model structure methods design modeling approaches to overcome the noise in the pseudo labels. Label refinement approaches include filtering methods used to remove unreliable pseudo labels and update methods aim to correct erroneous pseudo labels. Data enhancement methods artificially increase the diversity of the training data by performing various transformations, enhancements, or perturbations on the existing data. Lastly, optimization improvement methods concentrate on making adjustments to both the loss function and training strategy to ensure the effective use of both labeled and unlabeled datasets. While shown to be more flexible than supervised learning, these approaches still assume an initial set of labeled data, albeit significantly smaller than supervised methods. To enable learning from fully unlabeled datasets, our framework generates pseudo labels by using a model-free approach.

\subsection{Vision Foundation Models}

Recent advances in deep learning have seen the introduction of Vision Foundation Models (VFMs) \cite{awais2025foundation} that, unlike traditional models trained for a specific task, are large, versatile models capable of performing various vision tasks. Introduced by Kirillov et al. as one of the first VFMs, the Segment Anything Model (SAM) has since revolutionized the field of image analysis through its general scene understanding, advanced contextual reasoning, and capacity to perform several tasks. Designed using a Vision Transformer (ViT) \cite{dosovitskiy2020image} as its encoder, a prompt encoder that allows for incorporating human-provided input prompts (e.g., points, bounding boxes, masks), and a lightweight decoder, SAM was shown to achieve state-of-the-art results on numerous tasks while also containing exceptional zero-shot performance. Since its inception, researchers have proposed various advances to SAM including enhanced efficiency (FasterSAM \cite{zhang2023faster}), improved performance (HQ-SAM \cite{ke2023segment}), and the automation of its prompt encoder (AutoSAM \cite{shaharabany2023autosam}). Researchers have also adapted SAM to additional domains and tasks including medical images (MedSAM \cite{ma2024segment}), object tracking (TAM \cite{yang2023track}), and image captioning (CAT \cite{wang2023caption}). Given its widespread success, we acknowledge the benefit in adapting SAM for SR-CT; however, we recognize several challenges in this approach. First, as discovered in \cite{sahoo2024unveiling}, VFMs are still susceptible to hallucinations, which limits their direct application given the large differences between photographic and SR-CT datasets. Furthermore, while the work in \cite{ma2024segment} successfully adapted SAM to medical images, including CT, they used a large-scale dataset with 1,5720,263 image-mask pairs which cannot easily be replicated for SR-CT. Therefore, to limit the possibility of hallucinations and eliminate the need for manual labels, our framework follows traditional approaches by training a single-task model for each SR-CT dataset utilizing pseudo labels.

\begin{figure*}[t!]
\centering\includegraphics[width=1.0\linewidth]{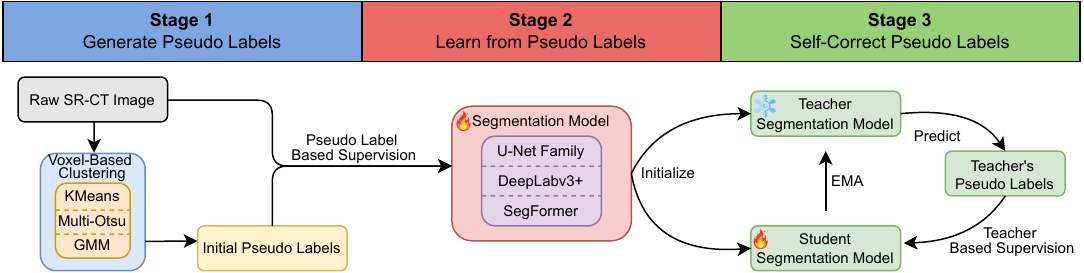}
\caption{Three stage approach for unsupervised semantic segmentation in synchrotron computed tomography (SR-CT) with self-correcting pseudo labels. In the first stage, we cluster voxel values to generate initial pseudo labels on unlabeled SR-CT images which are then used in the second stage to train a segmentation model. Finally, in the self-correction stage, we adapt the Unbiased Teacher\cite{liu2021unbiased} approach to refine the segmentation model.\label{fig:workflow}}
\end{figure*}

\section{Method}
\label{sec:method}

In this section, we introduce our framework for unsupervised semantic segmentation of SR-CT datasets with the overall approach illustrated in Figure \ref{fig:workflow}. In section \ref{sec:bn}, we formulate the problem and introduce the notation used. Section \ref{sec:plg} presents our approach to generating pseudo labels providing an initial semantic map. Then in Section \ref{sec:lpl}, we show how these pseudo labels are used to train a model allowing it to learn simple representations relating to structures with similar attenuation coefficients. Finally, we describe how these pseudo labels are then self-corrected using the Unbiased Teacher \cite{liu2021unbiased} approach resulting in a final model with an enhanced understanding of the data.    

\subsection{Problem Formulation}
\label{sec:bn}

Our goal is to address semantic segmentation of large, high-resolution SR-CT datasets in an unsupervised setting. We assume we are given an SR-CT sample containing $N$ images $\mathcal{X} = \lbrace x_i \rbrace _{i=1}^N$, in which we seek to produce a set of semantic masks $\mathcal{Y} = \lbrace y_i \rbrace _{i=1}^N$. Each $x_i \in \mathbb{R}^{H\times W \times 1}$ where $H$ and $W$ are image height and width, respectively, and $1$ represents the single channel measuring the amount of x-ray radiation absorbed by the material whereas $y_i \in \mathbb{R}^{H\times W \times K}$ with $K$ representing the number of classes. 

\subsection{Pseudo Label Generation}
\label{sec:plg}

As noted above, common approaches to generating pseudo labels involve utilizing a pretrained model to annotate unlabeled samples. Applied to medical CT, this setup can easily be replicated given the large-scale effort by researchers, physicians, and radiologists \cite{jiao2024learning} further facilitated by the high-similarity among different samples improving the reliability of the pseudo labels. However, SR-CT is applied to a wide variety of samples across scientific disciplines making this approach both impractical and unreliable motivating the use of a model-free approach. 

A key characteristic of CT, including SR-CT, is the interpretation of its voxel values \cite{buzug2008computed}. Unlike photographic images, where pixel values measure brightness and color at a specific location, CT voxels represent the amount of x-ray radiation absorbed by the material with its density, atomic composition, and compounds influencing the amount absorbed \cite{bezak2021johns}. Exploiting this property, we generate pseudo labels by clustering on voxel values making the assumption that structures with similar absorption values belong to the same category. While this assumption may not always hold in practice, it allows for an easy and scalable approach to generating pseudo labels. We leave the discussion of how we address noise in the pseudo labels to Section \ref{sec:scpl}.

Although there are numerous clustering algorithms, we use the KMeans algorithm \cite{Jin2010} given its widespread popularity, scalability, and ease of use, and provide a brief description here. Given a 2D SR-CT image $x_i$ with shape $H \times W \times 1$, we first flatten the image to get a 1D dimensional vector of $HW\times 1$ and define ${p_{i,l}}_{l=1}^{HW}$ to represent the $p$-th pixel in the $i$-th image. We drop $i$ below to simplify the notation. The KMeans algorithm is initialized based on minimizing its objective function

\begin{equation}\label{eq:1}
    J = \sum_{j=1}^K\sum_{p_l\in C_j}||p_{l}-\boldsymbol{\mu}_j||^2_2,
\end{equation}

\noindent where $J$ measures the total within-cluster sum of squares, $K$ is the number of classes/clusters, $C_j$ is the set of data points assigned to cluster $j$, $\boldsymbol{\mu}$ is the centroid of cluster $j$, and $||\cdot||_2^2$ is the squared Euclidean distance. After initialization, the $l$-th pixel is assigned the label based on

\begin{equation}\label{eq:2}
    c^* = \mbox{arg min}_{j\in \lbrace1,2,...,K\rbrace}||p_l - \boldsymbol{\mu}_j||_2^2, 
\end{equation}
which assigns pixels to the cluster whose centroid is closest to it in terms of Euclidean distance.

\subsection{Learning from Pseudo Labels}
\label{sec:lpl}

After generation, we train a segmentation model to learn from the initial pseudo labels. Doing so allows the model to learn simple relationships enabling it to identify structures with different absorption values. Learning from the pseudo labels follows standard approaches for training deep learning models for semantic segmentation tasks. However, for completeness, we provide the details of our initial training stage. Given an SR-CT image $x\in \mathbb{R}^{H \times W}$, the model outputs a per-pixel logit defined as 

\begin{equation}\label{eq:3}
    z = f_\theta(x) \in \mathbb{R}^{K \times H \times W}, 
\end{equation}

\noindent where $K$ is the number of clusters defined above. Class probabilities can be obtained using the softmax function applied independently at each pixel given by

\begin{equation}
\label{eq:4}
    p_{i,k} = \frac{\mbox{exp}(z_{i,k})}{\sum^K_{k'=1}\mbox{exp}(z_{i,k'})} \quad i\in \lbrace1,...,M\rbrace,
\end{equation}

\noindent where $M$ denotes the number of pixels ($H \times W$) and $p_{i,k}$ is the probability of class $k$ at pixel $i$. Supervised training is provided by the initial pseudo labels $c^*_i \in \lbrace 1,..K\rbrace$ with the one-hot target distribution defined as 

\begin{equation}\label{eq:5}
    q_{i,k} = 
        \begin{cases}
      1, & \text{if}\ k=c^*_i, \\
      0, & \text{otherwise}. 
    \end{cases}
\end{equation}

\noindent Optimizing $f_\theta$ uses the standard cross-entropy loss given by

\begin{equation}\label{eq:6}
    \mathcal{L}_{\text{CE}} = -\frac{1}{M}\sum^M_{i=1}\sum^K_{k=1}q_{i,k}\text{log}p_{i,k},
\end{equation}

\noindent with the optimal model being the one that achieves the minimum loss over the training dataset defined as

\begin{equation}\label{eq:7}
    \theta^* = \text{arg} \ \underset{\theta}   {\text{min}}\  \frac{1}{N}\sum^M_{j=1}\mathcal{L}_{CE}(f_\theta(x)^{(j)},c^{*(j)}).
\end{equation}

\noindent Finally, we note that the exact model architecture, loss function, and input strategy can be flexibly chosen to optimize performance and are further explored in Section \ref{sec:exp}. 

\subsection{Self-Correcting Pseudo Labels}
\label{sec:scpl}

\begin{figure}[t!]
\centering\includegraphics[width=1.0\linewidth]{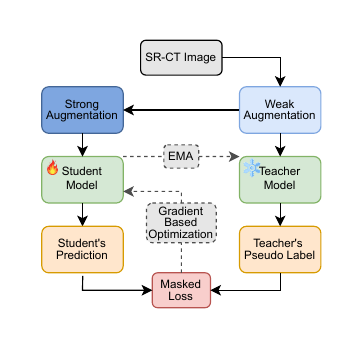}
\caption{Adaptation of the Unbiased Teacher \cite{liu2021unbiased} approach for unsupervised semantic segmentation. \label{fig:wsaug}}
\end{figure}

Although we have built a pseudo labeling process that eliminates the need for manual labels, allowing the model to learn in an unsupervised fashion, we have assumed that significant differences in absorption values are sufficient to identify and categorize different structures within the volume. However, in practice, this assumption may not always hold as SR-CT is regularly corrupted by varying levels of background noise and imaging artifacts \cite{buzug2008computed} which systematically biases voxel values effectively degrading the reliability of the cluster-based pseudo labels. Additionally, these pseudo labels only capture features relating to differences in absorption values which may limit model generalization as the model lacks an understanding of other key features including shape, texture, edges, etc. 

To improve over such noisy pseudo labels and instill a more holistic understanding of the data into the model, we adapt the Unbiased Teacher \cite{liu2021unbiased} learning paradigm for unsupervised semantic segmentation. Before introducing our adaption of the Unbiased Teacher approach for unsupervised semantic segmentation, we first briefly describe the mutual-learning process in \cite{liu2021unbiased}. Originally developed for semi-supervised object detection, the Unbiased Teacher method utilizes a student-teacher mutual learning approach that aims at evolving both models via a mutual learning mechanism. By having the teacher generate pseudo labels to train the student, the student is able to update its knowledge back to the teacher effectively improving the next round of pseudo labels allowing both models to evolve jointly and continuously improve detection accuracy. Key to the approach is the use of augmented images during mutual learning. To ensure the diversity of student models, strong augmentations are used; whereas weak augmentations are used as input to the teacher to generate reliable pseudo labels. While the student model is updated via gradient descent, the teacher model is updated as an exponential moving average (EMA) of the student. In the semi-supervised setting, the student receives supervision from the teacher's pseudo labels and the images with ground-truth labels. Additionally, to obtain more reliable supervision from the teacher's predictions, a confidence threshold $\delta$ is set to filter low-confidence predicted bounding boxes. 

Adapting the Unbiased Teacher approach for unsupervised semantic segmentation requires two changes with Figure \ref{fig:wsaug} illustrating the version used in this work. First, unlike \cite{liu2021unbiased}, we no longer have a set of images with ground-truth labels originally used as one of the sources of supervision for the student. Rather, we have the pseudo labels generated in Section \ref{sec:plg}. Although these could be used to supervise the student, similar to our initial learning stage, any noise in the pseudo labels will provide contradictory learning signals to the student ultimately hampering its ability to evolve and improve segmentation accuracy. Therefore, we drop this source of supervision in our framework. Second, \cite{liu2021unbiased} sets a confidence threshold to filter low-confidence bounding boxes which needs to be adapted for semantic segmentation. This corresponds to masking low-confident pixels which can be defined as 

\begin{equation}\label{eq:8}
    m_{i} = 
        \begin{cases}
      1, & \text{if}\ \underset{k}   {\text{max}}\ p^T_{i,k} > \delta, \\
      0, & \text{otherwise},
    \end{cases}
\end{equation}

\noindent where $p^T_{i,k}$ is the probability of class $k$ at pixel $i$ predicted by the teacher model $f^T_{\theta}$. Optimizing the student model, $f^S_\theta$, using the cross-entropy loss is defined similar to Equation \ref{eq:6} but includes the pixel-level mask $m_{i}$

\begin{equation}\label{eq:9}
    \mathcal{L}_{\text{MASK-CE}} = -\frac{1}{\sum_im_i}\sum^M_{i=1}m_{i}\sum^K_{k=1}q^T_{i,k}\text{log}p^S_{i,k},
\end{equation}
where $p^S_{i,k}$ is the probability of class $k$ at pixel $i$ predicted by the student model and $q^T_{i,k}$ is one-hot target distribution provided by the teacher model. We note that rather than normalize by the total number of pixels $M$, we normalize by the number of high-confident pixels $\sum_im_i$ to maintain a stable loss during training given the number of pixels per update.  Finally, we follow \cite{liu2021unbiased} in updating the teacher's weights by applying EMA defined as

\begin{equation}\label{eq:10}
    \theta_T \leftarrow \ \alpha \theta_T + (1-\alpha)\theta_S,
\end{equation}

\noindent where $\alpha \in (0,1)$ is the momentum coefficient.

\section{Experiments}
In this section, we thoroughly evaluate our proposed framework on various real-world SR-CT datasets. Additionally, we determine the flexibility and robustness of our framework to different design choices for each of the three stages.

\label{sec:exp}
\subsection{Experimental Setup}
\noindent\textbf{Datasets.} We evaluate our framework on the following three real-world SR-CT datasets collected at the APS beamline 1-ID at Argonne National Laboratory:

\noindent\textbf{Magnesium Crystal} \cite{plumb2023dark} was cleaved from a single-crystal block of NaMnO$_2$ quantum material with the primary investigation being into the location and structure of defects in the lattice of the crystalline material at the mesoscale impacting its controversial low-temperature phase transformation and magnetic behavior. The defect of interest in the room-temperature imaging study is a tetragonal intergrowth phase of Mn$_3$O$_4$ into the monoclinic $\alpha$-NaMnO$_2$ primary phase that could be identified and mapped. During acquisition, the scans were acquired using a 2.15 mm (width) $\times$ 1.0 mm (height) beam while rotating the apparatus over $360\degree$ at $0.2\degree$ angular increments with 0.5 second exposure time. This procedure produced 1800 projection images of $1520 \times 1200$ pixels that were reconstructed into a 3D volume with 1200 2D slices of $1520 \times 1520$ pixels with 1.17$\mu m$ voxel spatial resolution.

\noindent\textbf{Silica Sand} \cite{imseeh2020influence} contains 2705 silica sand particles loaded into a mechanical compression apparatus. This dataset was used to investigate the influence of quartz crystal structure on the constituitive response of silica sand particles using synchrotron micro-CT and high-energy diffraction microscopy. During acquisition, the scans were acquired using a monochromatic 72 keV, 1.6 mm (width) $\times$ 1.2 mm (height) x-ray beam while rotating the apparatus over $360\degree$ at $0.2\degree$ angular increments with 0.28 second exposure time. This procedure produced 1800 projection images of $1520 \times 1200$ pixels that were reconstructed into a 3D volume with 1200 2D slices of $1520 \times 1520$ pixels with 0.98$\mu m$ voxel spatial resolution.

\noindent\textbf{Ceramic Prism} contains a solid AlON square cross-section prismatic sample, in which a pre-initiated crack morphology and propagation was followed with sequential tomography scans. The contrast of the thin crack in a highly transparent material was in the same range as the experimental noise level. The scans were acquired with a monochromatic 65 keV, 2.1 mm (width) $\times$ 1.0 mm (height) x-ray beam, and similar to the previous data set, using $360\degree$ full rotation at $0.2\degree$ angular increments with 0.08 second exposure time. Like above, 1800 projection images of $1520 \times 1200$ pixels that were reconstructed into a 3D volume with 1200 2D slices of $1520 \times 1520$ pixels with BLANK$\mu m$ voxel spatial resolution.

\subsection{Implementation Details}

The training environment used in this study consists of 4 NVIDIA Tesla V100 GPUs with 32GBs each. Each of the models considered uses group normalization \cite{wu2018group} and dropout layers to reduce the risk of overfitting and swaps ReLU for PReLU as the nonlinearity layer implemented using PyTorch 2.0.1\cite{paszke2019pytorch}. Unless otherwise stated, the models are trained for 200 epochs in stage 2 before training for an additional 200 epochs in stage 3 using the Adam optimizer\cite{kingma2014} with learning rate $1e-03$ and batch size of 16. For the data augmentation, weak augmentations follow geometrical transformations including 90$\degree$, 180$\degree$, and 270$\degree$ rotations and vertical, horizontal, and diagonal flips. Strong augmentations follow photometric augmentations including random gamma correction, color jitter, contrast limited adaptive histogram equalization (CLAHE), and random changes to the brightness and contrast of the image. Both the weak and strong augmentations are implemented using the Albumentations 2.0.8 library \cite{info11020125}. Before the augmentations, we scale the images between $[0,1]$ and use a randomly cropped image of size $512 \times 512$ as the input. Finally, during the self-correction stage, we set $\delta=0.5$ and $\alpha=0.99$. 

\begin{figure}[b!]
\centering\includegraphics[width=1.0\linewidth]{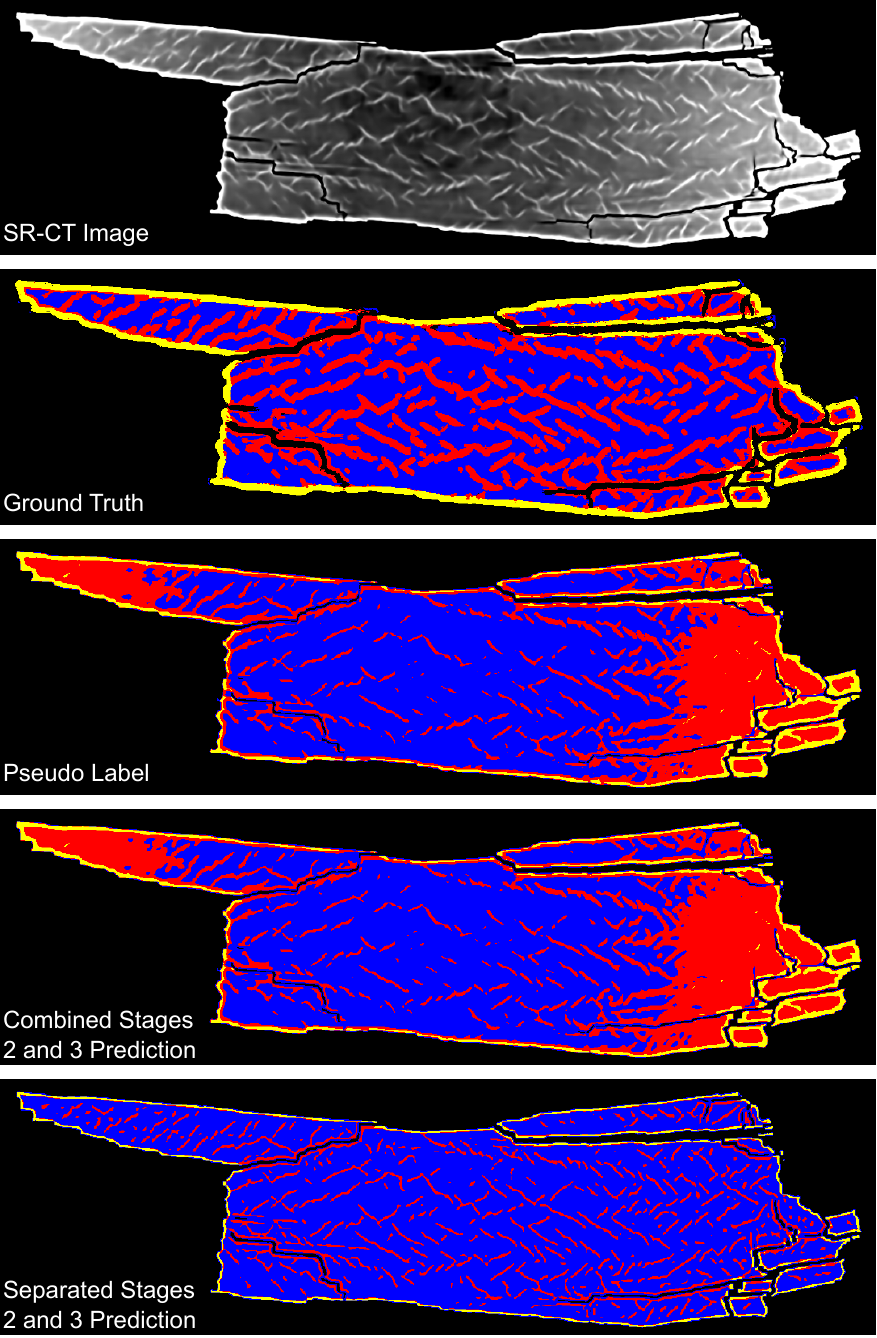}
\caption{Evaluating the necessity of the self-correction stage. \label{fig:exp1}}
\end{figure}

\subsection{Evaluation Metrics}
Given the high manual cost of labeling 1200 high-resolution slices, we manually label 15 representative slices of the magnesium dataset to quantitatively evaluate the performance of our framework using both pixel accuracy and mean intersection-over-union (mIoU) metrics. Additionally, we exclude the background during evaluation as this region is not relevant from an investigator's viewpoint. We use the sand and ceramic sample to evaluate the generalizability of our framework and rely on visual examination to determine performance. 

\subsection{Evaluating the Necessity of the Self-Correction Stage}

We first evaluate the necessity of the self-correction stage. Following \cite{liu2021unbiased}, our approach uses weak and strong data augmentation strategies in stage 3 to enable the model to evolve and improve segmentation accuracy. However, this augmentation strategy can be combined with the initial pseudo label learning stage potentially eliminating the need for two separate training stages. Therefore, we compare two approaches where one trains a model using just the initial pseudo labels with 50\% of the batch containing strong augmentations and the remaining half containing no augmentation. The second approach follows our proposed framework. 

Figure \ref{fig:exp1} shows an example SR-CT image from the magnesium dataset cropped to the sample including the ground truth, pseudo label, and model predictions both with and without the self-correction stage when using a simple U-Net\cite{ronneberger2015u}. First, from the SR-CT image, we can identify the location of the intergrowth phases in the NaMnO$_2$ sample as the faint white streaks inside the material which are highlighted red in the ground truth label. The other classes include the background (black), sample (blue), and sample boundary (yellow). The pseudo labels used for training are capable of identifying these structures but abruptly changes classes as the contrast of the sample shifts given slight changes in the vanadium content. By combining the pseudo labels and strong augmentations during training, we observe that the model is unable to evolve and predicts an image identical to the pseudo label. However, by separating the initial learning and self-correction stages, we can see that the model becomes robust to slight changes in contrast and enhances the connectedness of the intergrowth phases. Quantitatively, the model that combines both stages reports pixel accuracy and mIoU as 66.94\% and 46.13\%, respectively. When separating the training stages, pixel accuracy jumps to 67.75\% whereas mIoU drops to 40.30\%. We note that this contradiction between the quantitative and qualitative metrics reflects the predicted intergrowth phases and sample boundary (yellow) in the self-correction approach being much thinner than those identified in the ground truth image. Although the approaches appear to be similar quantitatively, qualitatively, the results indicate that separating the initial learning and self-correction stages is necessary to improve segmentation accuracy. 

\begin{figure}[b!]
\centering\includegraphics[width=1.0\linewidth]{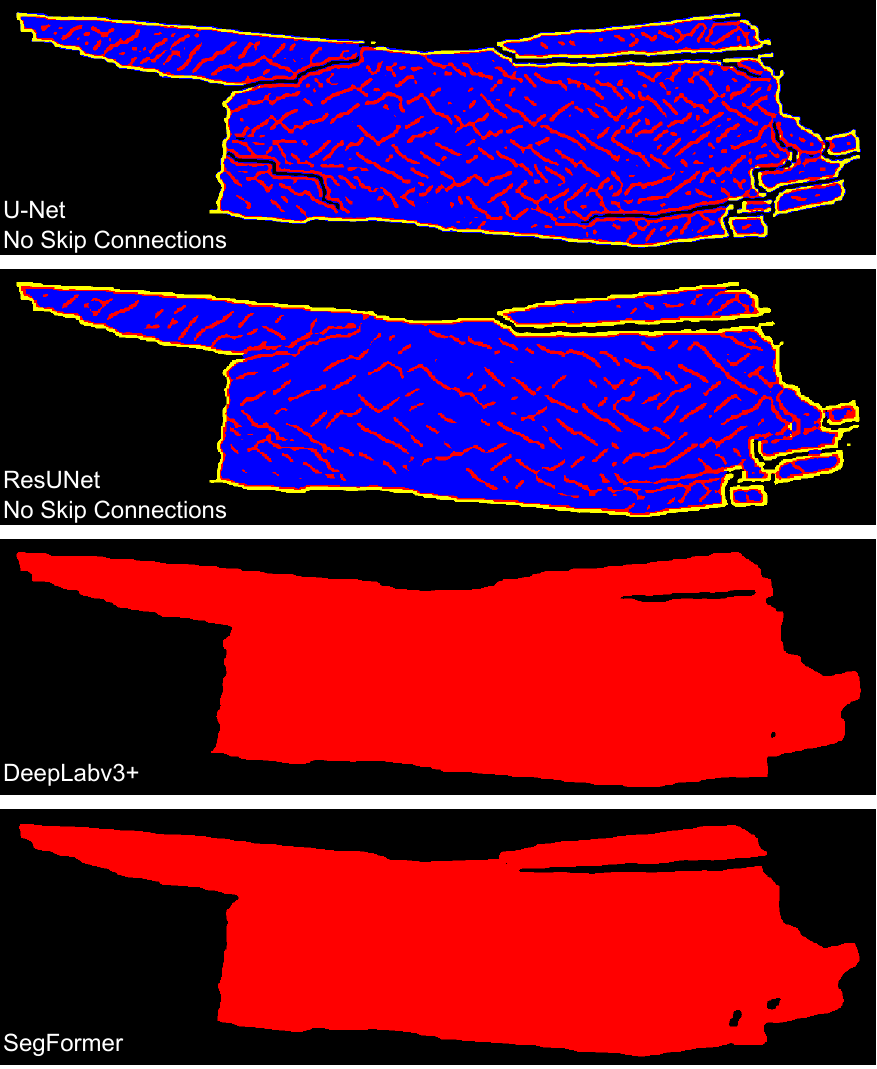}
\caption{Stage 3 model predictions for U-Net and ResUNet both with no skip connections, DeepLabv3+, and SegFormer. \label{fig:exp2}}
\end{figure}

\subsection{Comparing Different Semantic Segmentation Models}

We now compare the performance when using different deep learning segmentation models. While U-Net remains a popular choice and competitive baseline, numerous extensions have been proposed including the use of enhanced skip-connections (UNet++\cite{zhou2018unet++}, UNet3+\cite{huang2020unet}), residual units (ResUNet\cite{zhang2018road}, ResUNet++\cite{jha2019resunet++}), and attention gates (Attention U-Net\cite{oktay2018attention}). Furthermore, we explore these models without their skip connections as these components enable transferring high-resolution spatial information from the encoder to the decoder which may hamper model performance when using noisy pseudo labels. Finally, we also include DeepLabv3+\cite{chen2018encoder} which combines the spatial pyramid pooling module with the encoder-decoder structure as well as SegFormer\cite{xie2021segformer}, a purely transformer-based model. For a fair comparison, all models are implemented to contain approximately 2 million trainable parameters. 

\begin{table}[h]
\begin{adjustbox}{width=\columnwidth}
\begin{threeparttable}
\begin{tabular}{@{}lcccc@{}}
\toprule
                        & \multicolumn{2}{c}{Stage 2}                                   & \multicolumn{2}{c}{Stage 3}                                   \\ \midrule
Model                   & \multicolumn{1}{l}{Accuracy} & \multicolumn{1}{l}{mIoU} & \multicolumn{1}{l}{Accuracy} & \multicolumn{1}{l}{mIoU} \\ \midrule
U-Net\cite{ronneberger2015u}                   & 66.89                             & 46.08                    & 67.75                              & 40.30                     \\
U-Net\tnote{*}         & 66.57                              & 45.82                    & \textbf{73.43}                             & \textbf{51.01}                   \\
UNet++\cite{zhou2018unet++}                  & 67.26                              & 45.99                    & 55.18                              & 19.35                    \\
UNet3+\cite{huang2020unet}                  & 65.53                              & 43.41                    & 61.60                               & 34.46                    \\
ResUNet\cite{zhang2018road}                 & 65.23                              & 42.74                    & 60.46                              & 26.66                    \\
ResUNet\tnote{*}          & \textbf{69.37}                              & \textbf{48.30}                     & 73.15                              & 50.31                    \\
ResUNet++\cite{jha2019resunet++}               & 66.72                              & 44.91                    & 65.76                              & 44.69                    \\
ResUNet++\tnote{*}       & 68.60                               & 47.81                    & 62.21                              & 36.48                    \\
Att. U-Net \cite{oktay2018attention}         & 68.00                                 & 45.50                     & 58.51                              & 26.84                    \\
Att. U-Net\tnote{*}  & 66.35                              & 44.12                    & 53.68                              & 18.90                     \\
DeepLabv3+\cite{chen2018encoder}              & 65.37                              & 43.28                    & 54.91                              & 17.38                    \\
SegFormer\cite{xie2021segformer}               & 66.89                              & 46.73                    & 54.90                               & 18.11           \\ \bottomrule       
\end{tabular}
\begin{tablenotes}
    \item[*] indicates model has no skip connections.
\end{tablenotes}
\end{threeparttable}
\end{adjustbox}
\caption{Stage 2 and 3 pixel accuracy and mIoU metrics for different segmentation models.}\label{tab:e2}
\end{table}

Table \ref{tab:e2} shows the quantitative results for the different models considered. We report both the stage 2 and stage 3 results to understand the changes that exist within the models between the two stages as well as across the models. First, we can see that all models report consistent metrics in stage 2 with minor differences between the lowest and highest value indicating that all models are capable of learning from the initial pseudo labels. Second, unlike stage 2, stage 3 shows significant variations in the reported metrics, especially mIoU. Although UNet++, DeepLabv3+, and SegFormer performed adequately in stage 2, these models show sharp drops in their mIoU values in stage 3. This is further verified in Figure \ref{fig:exp2} which shows both the DeepLabv3+ and SegFormer models are no longer able to separate the details within the sample; instead, they collapse into predicting the sample as a single class. Interestingly, both Table \ref{tab:e2} and Figure \ref{fig:exp2} suggest that a simple U-Net model with no skip connections performs the best with the identified intergrowth phases better aligning with the ground truth image compared to those predicted by the standard U-Net in Figure \ref{fig:exp1}. The remaining experiments are performed using the U-Net model with no skip connections.

\subsection{Evaluating the Effects of Different Loss Functions}
In addition to different segmentation models, we explore the impact of different loss functions. Given the popularity in using pseudo labels for training, researchers have developed various noisy label learning strategies including the use of robust loss functions such as generalized cross entropy (GCE) \cite{zhang2018generalized} and symmetric cross entropy (SCE) \cite{wang2019symmetric} that have demonstrated improved performance. We also consider confidence calibration techniques designed to address the issue of over-confidence common in deep learning models \cite{guo2017calibration} via the use of label smoothing \cite{szegedy2016rethinking} and bootstrapping \cite{reed2014training}. Finally, we include the focal loss as it was demonstrated in \cite{ma2020normalized} to be effective in the presence of noisy labels.  

\begin{table}[h]
\begin{adjustbox}{width=\columnwidth}
\begin{tabular}{@{}lcccc@{}}
\toprule
                      & \multicolumn{2}{c}{Stage 2} & \multicolumn{2}{c}{Stage 3} \\ \midrule
Loss Function         & Accuracy       & mIoU       & Accuracy       & mIoU       \\ \midrule
Cross Entropy         & 66.57          & 45.82      & 73.43          & 51.01      \\
Generalized CE\cite{zhang2018generalized}   & 66.62          & 45.69      & 69.29          & 43.44      \\
Symmetric CE\cite{wang2019symmetric}   & 67.11          & 45.21      & 68.55          & 41.17      \\
Label Smoothing\cite{szegedy2016rethinking} & 67.16          & 45.18      & \textbf{75.47}          & \textbf{52.65}      \\
Bootstrapping\cite{reed2014training}   & 66.54          & 44.17      & 74.23          & 51.75       \\
Focal\cite{ross2017focal}                 & \textbf{67.79}          & \textbf{46.33 }     & 69.99          & 43.08      \\ \bottomrule
\end{tabular}
\end{adjustbox}
\caption{Stage 2 and 3 pixel accuracy and mIoU metrics for different loss functions.}\label{tab:e3}
\end{table}

\begin{figure}[t!]
\centering\includegraphics[width=1.0\linewidth]{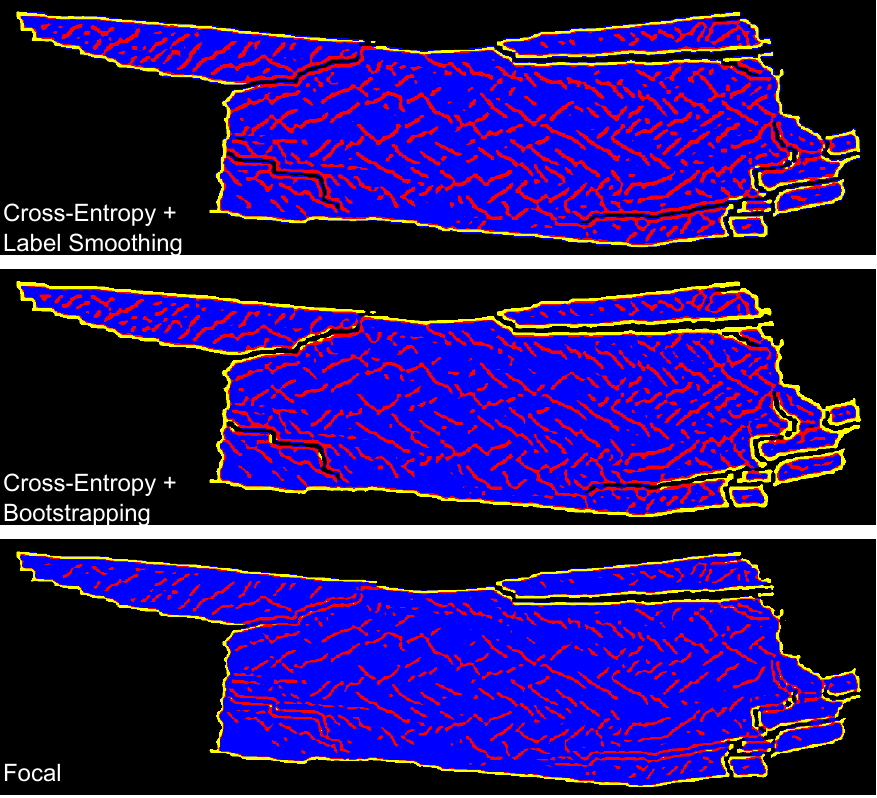}
\caption{Stage 3 model predictions when using cross entropy loss with label smoothing and bootstrapping and focal loss. \label{fig:e3}}
\end{figure}

Table \ref{tab:e3} shows the quantitative results for the different loss functions. First, like we saw above, each of the loss functions perform similarly in stage 2 with some larger differences seen in stage 3. Second, although the focal loss performs the best in stage 2, it produces the second worst results in stage 3 as seen by its mIoU value. Finally, we observe that the confidence calibration techniques clearly outperform the robust loss function methods. Given that stage 3 only uses the teacher's predictions as supervisory signal to the student, these calibration techniques enable the model to become more reliable in its predictions, boosting performance as low-confidence pixels are masked during optimization. Figure \ref{fig:e3} shows the predicted sample when training using label smoothing, bootstrapping, and the focal loss. As the quantitative results suggest, the segmentations are nearly identical to that seen in Figure \ref{fig:exp2}. While the focal loss is able to improve over the change in material content, some of the small gaps are blended into the sample. The remaining experiments are conducted using the cross entropy loss with label smoothing. 

\subsection{Evaluating the Effects of Different Input Strategies}

\begin{table}[b!]
\begin{adjustbox}{width=\columnwidth}
\begin{tabular}{@{}lcccc@{}}
\toprule
             & \multicolumn{2}{c}{Stage 2} & \multicolumn{2}{c}{Stage 3} \\ \midrule
Adjacent Slices & Accuracy       & mIoU       & Accuracy       & mIoU       \\ \midrule
1            & 67.16          & 45.18      & 75.47          & 52.65      \\
3            & 67.03          & 45.62      & 75.92          & 53.22      \\
5            & 65.42          & 44.17      & 73.66          & 52.08      \\
7            & \textbf{67.40}          & \textbf{46.11}      & \textbf{76.43}          & \textbf{54.41}      \\
9            & 66.53          & 44.88      & 75.07          & 53.58      \\ \bottomrule
\end{tabular}
\end{adjustbox}
\caption{Stage 2 and 3 pixel accuracy and mIoU metrics for different number of input slices.}\label{tab:e4}
\end{table}

Up to this point, we have processed the SR-CT datasets as a collection of independent 2D images. However, these volumes are inherently 3D implying that structures that exist in a given image most likely exist across a range of neighboring slices. Although 3D deep learning models have been shown to outperform their 2D counterparts \cite{hatamizadeh2022unetr}, these approaches are computationally costly and require large GPU memory restricting their use in practice. Originally introduced in \cite{ziabari20182}, the 2.5D approach stacks a group of adjacent 2D slices together and treats them as a multi-channel input in order to capture some 3D contextual information without the full memory overhead of 3D approaches. Following \cite{ziabari20182}, we evaluate the impact this approach has on model performance.

\begin{figure}[h!]
\centering\includegraphics[width=1.0\linewidth]{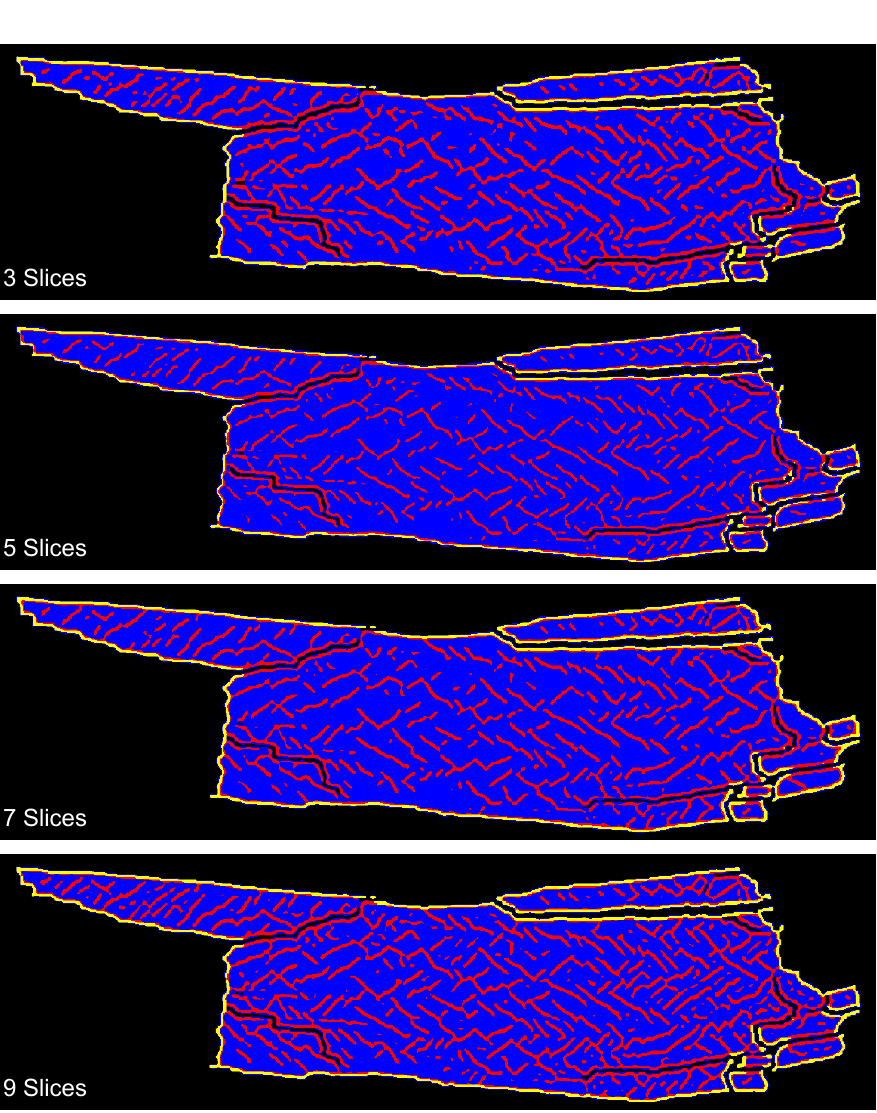}
\caption{Stage 3 model predictions when using 3, 5, 7, and 9 adjacent slices as input following the 2.5D approach. \label{fig:e4}}
\end{figure}

Table \ref{tab:e4} shows the quantitative results when inputting 1, 3, 5, 7, and 9 slices into the model. Given the target image is the middle slice, the odd number allows the model to incorporate an equal amount of information from both neighboring sides. To ensure consistency across the input stack of images, both the weak and strong augmentations are replicated across all slices. From Table \ref{tab:e4}, we can see only minor variations exist between stage 2 and 3. The results indicate that using 7 slices as input gives the best performance although only slightly above the others. Figure \ref{fig:e4} illustrates a similar conclusion, yet we remark that the predicted intergrowth phases when using 3 and 9 slices are slightly fuller than when using 5 slices. The remaining experiments use a stack of 7 slices as input to the model.  

\subsection{Evaluating the Effects of Different Pseudo Labeling Strategies}
In this section, we explore the use of different pseudo labeling strategies to generate the initial labels used to train the model in stage 2. Given our approach to generating labels involves clustering on the voxel values, numerous strategies can be employed to achieve this. In this work, we compare the KMeans approach, used in previous experiments, to the Multi-Otsu\cite{liao2001fast} and Gaussian Mixture Model (GMM) \cite{xu2005survey} approaches. Multi-Otsu clusters the data similar to KMeans by maximizing the between class variance of voxel intensity whereas GMM fits a Gaussian mixture model to the data. Our implementation of GMM allows each class to have its own general covariance matrix. 

\begin{table}[h]
\begin{adjustbox}{width=\columnwidth}
\begin{tabular}{@{}lccccc@{}}
\toprule
           & \multicolumn{1}{c}{Stage 1}          & \multicolumn{2}{c}{Stage 2} & \multicolumn{2}{c}{Stage 3} \\ \midrule
Method      & \multicolumn{1}{c}{Time} & Accuracy       & mIoU       & Accuracy       & mIoU       \\ \midrule
KMeans     & \textbf{21.99}                         & \textbf{67.40}          & \textbf{46.11}      & \textbf{76.43}          & \textbf{54.41}      \\
Multi-Otsu & 27.13                         & 66.49          & 45.83      & 74.98           & 52.67      \\
GMMs       & 33.39                         & 67.44          & 48.06      & 73.62          & 51.13      \\ \bottomrule
\end{tabular}
\end{adjustbox}
\caption{Stage 2 and 3 pixel accuracy and mIoU metrics when using KMeans, Multi-Otsu, and Gaussian Mixture Models (GMMs) to generate the pseudo labels in stage 1. The wall time to generate the pseudo labels, in seconds, is also reported. }\label{tab:e5}
\end{table}

\begin{figure}[b!]
\centering\includegraphics[width=1.0\linewidth]{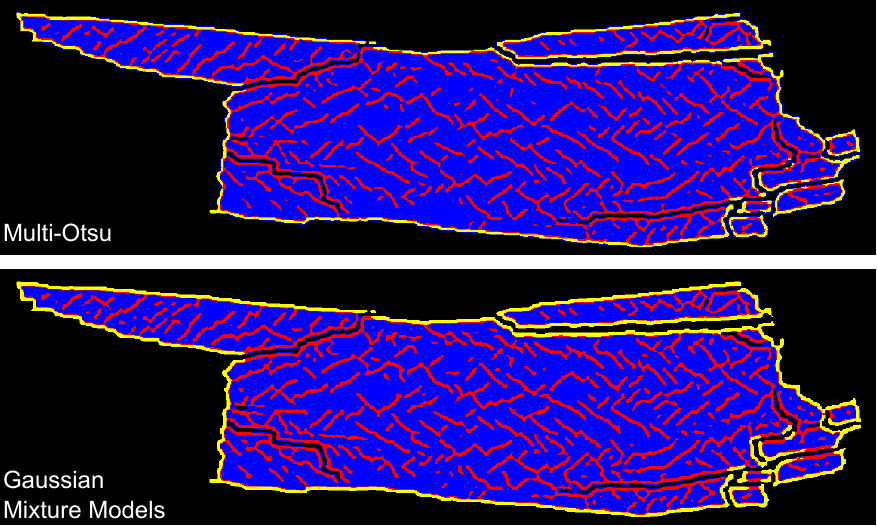}
\caption{Stage 3 model predictions when using Multi-Otsu and Gaussian Mixture Models to generate the initial pseudo labels for stage 2\label{fig:ep5}}
\end{figure}

Table \ref{tab:e5} shows the quantitative results between the different pseudo labeling strategies. We also include the wall time, in seconds, to generate the pseudo labels in stage 1. First, we note that each approach performs adequately well across both stages with KMeans reporting the best pixel accuracy and mIoU metrics in stage 3. Additionally, KMeans generates pseudo labels in 21.99 seconds, a 23.37\% and 33.14\% reduction in time over the Multi-Otsu and GMM methods, respectively. Figure \ref{fig:ep5} depicts the model predictions for the Multi-Otsu and GMM approaches. Overall, these predictions contain only slight differences between each other and the KMeans approach seen in Figure \ref{fig:e4}. The remaining experiments uses the pseudo labels generated from KMeans.

\subsection{Evaluating Sensitivity to Duration of Training in Stage 2}
An important consideration in our framework is how long to train the model in stage 2. Trained for too long, the model may overfit to the noisy pseudo labels and be unable to evolve in stage 3. However, if the model is under trained in stage 2, it may not fully capture the necessary relationships and collapse during stage 3 as it is unable to handle the strong augmentations. Furthermore, under trained models may produce the correct prediction but assign it a low confidence value ultimately hurting the model's ability to evolve as these correct, low confident predictions are masked in stage 3. Therefore, we explore the impact training in stage 2 has on segmentation accuracy in stage 3. To do this, we train the model for 1000 epochs in stage 2 and save the best model after every hundredth epoch. These models are then trained for an additional 200 epochs in stage 3.

\begin{table}[h]
\begin{adjustbox}{width=\columnwidth}
\begin{tabular}{@{}lccccc@{}}
\toprule
              & \multicolumn{3}{c}{Stage 2} & \multicolumn{2}{c}{Stage 3} \\ \midrule
Epoch & Accuracy       & mIoU & Confidence   & Accuracy       & mIoU       \\ \midrule
100           & 66.54          & 45.35 &  82.36    & 74.36          & 51.20       \\
200      &     67.40          & 46.11  &  86.50    & \textbf{76.43}          & \textbf{54.41}       \\
300           & \textbf{67.81}          & \textbf{46.68} &  87.87    & 68.77          & 40.82      \\
400           & 67.47          & 45.73 &  87.08    & 69.32          & 41.60       \\
500           & 67.27          & 46.26 &  87.68    & 68.00             & 39.97      \\
600           & 67.42          & 46.27 &  88.42    & 69.07          & 40.62      \\
700           & 67.34          & 46.10 &  87.74     & 68.26          & 40.49      \\
800           & 67.22          & 45.94 &  \textbf{89.58}    & 68.73          & 40.26      \\
900           & 67.34          & 46.45 &  89.56    & 67.03          & 37.91      \\
1000          & 67.31          & 46.15 &  89.39    & 67.88          & 39.03      \\ \bottomrule
\end{tabular}
\end{adjustbox}
\caption{Stage 2 and 3 pixel accuracy and mIoU metrics for different stopping points in stage 2. Average model confidence in stage 2 is also reported.}\label{tab:e6}
\end{table}

\begin{figure}[t!]
\centering\includegraphics[width=1.0\linewidth]{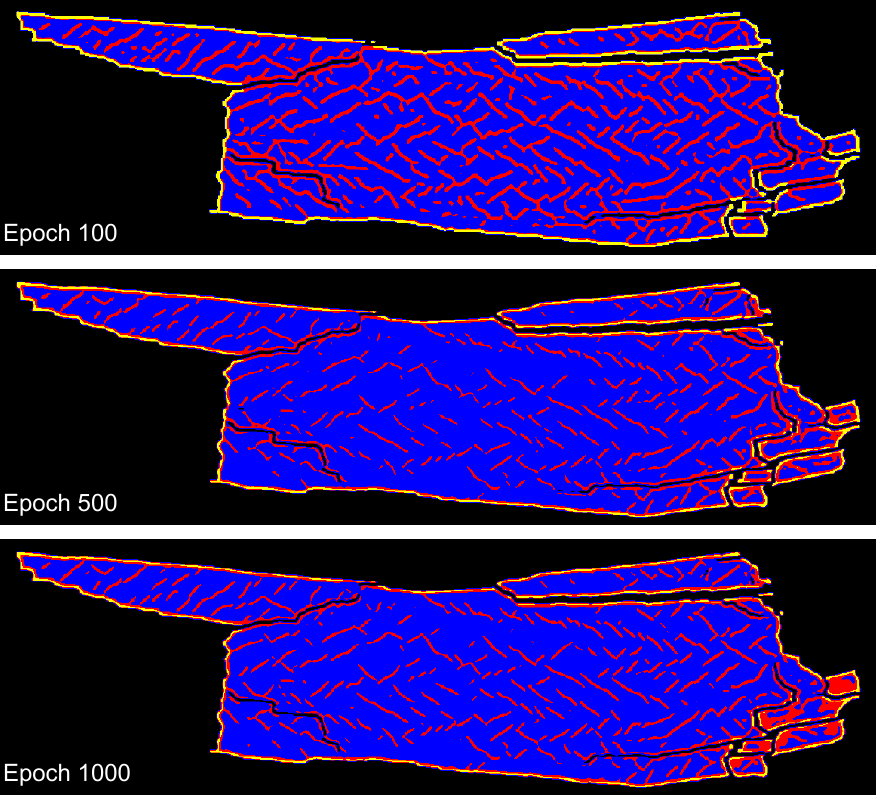}
\caption{Stage 3 model predictions when training for 100, 500, and 1000 epochs in stage 2. \label{fig:e6}}
\end{figure}

Table \ref{tab:e5} shows the quantitative results between the different stopping points in stage 2. We also include model confidence given the role it plays in stage 3. First, as expected, the model's performance continuously improves in the first few hundred epochs in stage 2 as seen in the pixel accuracy. Second, model confidence jumps sharply between epoch 100 and 200 before slowly trending upwards with a final spike seen at epoch 800. Finally, unlike stage 2, stage 3 shows a steady decline in mIoU as models are trained longer in stage 2 reaching its lowest value of 37.91 in epoch 900, a 30.33\% drop from 54.41 at epoch 200. Figure \ref{fig:e6} shows the stage 3 model predictions when trained for 100, 500, and 1000 epochs in stage 2. We can see the drop in mIoU correlates to the model predicting the intergrowth phases as extremely thin, with them no longer as connected as seen in the model trained for 100 epochs. Furthermore, the model is unable to correct some regions as seen in the lower-right portion of the sample.

\subsection{Evaluating the Effects on the Number of Classes}

\begin{figure}[t!]
     \centering
     \begin{subfigure}[b]{\linewidth}
         \centering
         \includegraphics[width=1.0\linewidth]{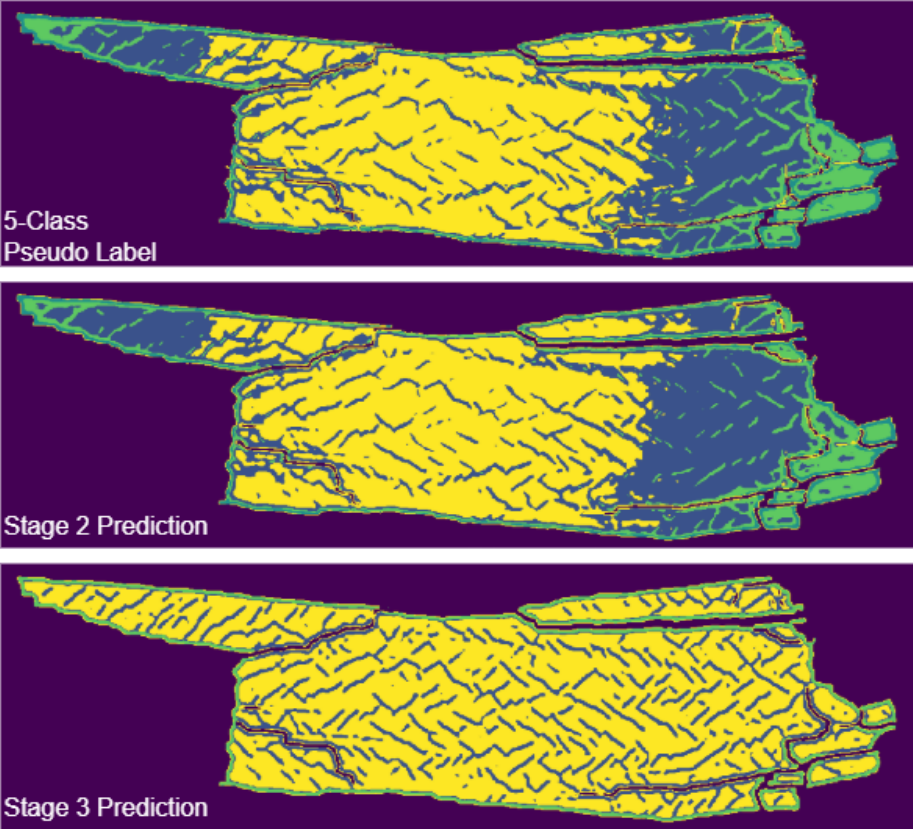}
         \caption{Results when using overly segmented pseudo label consisting of 5 classes including pseudo label, and stage 2 and 3 predictions. 
         }
     \label{fig:e7_5C_preds}
     \end{subfigure}
     \hfill
     \begin{subfigure}[b]{\linewidth}
         \centering
         \includegraphics[width=1.\linewidth]{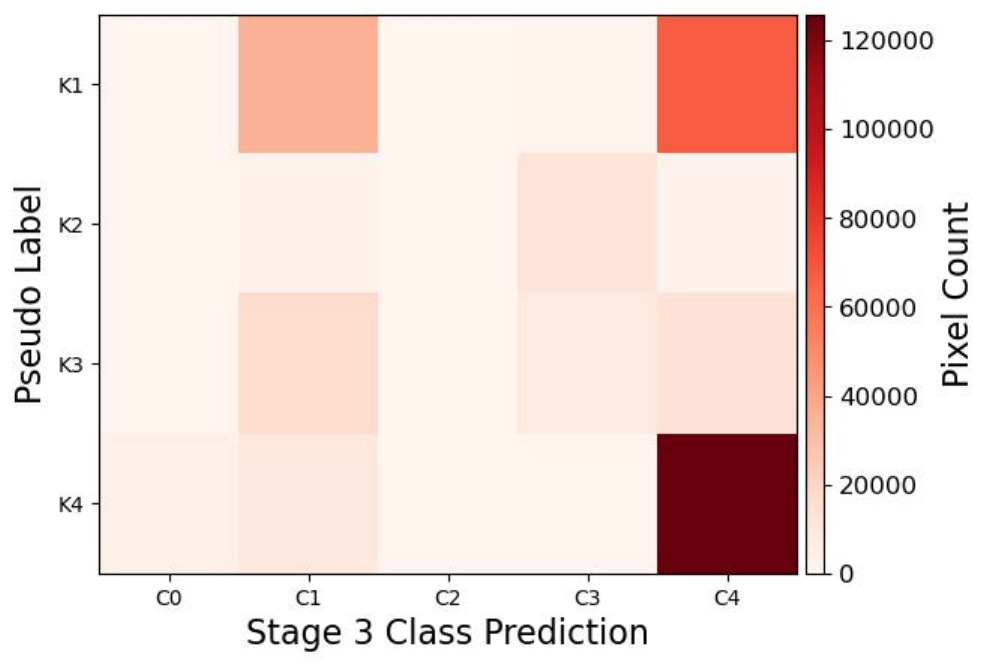}
         \caption{Confusion matrix showing the change in class assignment between the pseudo label and stage 3 when using 5 classes. Each cell represents the number of pixels that started in cluster K and ended in class C.}
     \label{fig:e7_5C_cm}
     \end{subfigure}
     \caption{Qualitative results for over-segmented SR-CT image consisting of 5 classes.}\label{fig:e7_5C}
\end{figure}

In this section, we evaluate the impact the number of classes has on segmentation accuracy. Until now, we have assumed that the number of classes is known prior to training. While researchers may know the compositional makeup of their sample, allowing the number of classes to reflect this intuition, in some instances this may not be known a priori. Although the elbow method \cite{yuan2019research} is a popular approach to identifying the optimal number of clusters when using KMeans, its usage in identifying meaningful structures in SR-CT datasets may not always be appropriate warranting an evaluation on how sensitive the framework is to the number of classes generated in stage 1. Given that segmentation models are architecturally upper-bounded by the number of classes, we only examine when the number of classes are over-estimated as models can collapse redundant regions into a single class but cannot create more classes than its design allows. In the first example, we set the number of classes to just one above optimal, 5, reflecting when the number of classes is misspecified but close to optimal. In the second example, we set the number of classes to 10, representing an extreme case where the number of classes is grossly over-estimated resulting in a highly fragmented segmentation. 

Figure \ref{fig:e7_5C_preds} depicts the 5-class pseudo label for the SR-CT slice from the magnesium dataset. We also include the model predictions after stage 2 and 3. First, we can see the main difference in the pseudo label compared to that used in previous experiments is the split in class assignment between the intergrowth phases in the center of the sample compared to those at the edges reflecting the change in vanadium content. Second, we observe that the model is able to handle this additional split during the initial training stage as seen by the model prediction closely matching the pseudo label. Although this result may indicate the model overfitting to slight variations in contrast, we observe that the prediction in stage 3 shows no substantial degradation compared to previous experiments. The self-correction stage is able to combine the intergrowth phase and sample background into a single class and drops one of the classes altogether from the final segmentation. To understand the change in class assignment between the initial pseudo label and the final segmentation, Figure \ref{fig:e7_5C_cm} shows a confusion matrix where each cell represents the number of pixels that started in cluster K and ended in class C, excluding the background cluster K0. We can see that the intergrowth phases, split between cluster K1 and K3, are consolidated into class C1 after self-correction. A similar trend can be seen in the sample background being combined into class C4. Finally, we remark that column C2 is empty indicating that after stage 3, the model assigns no pixels to this class. Instead, these pixels are spread among classes C1, C3, and C4 shown in row K3.  

\begin{figure}[b!]
     \centering
     \begin{subfigure}[b]{\linewidth}
         \centering
         \includegraphics[width=1.0\linewidth]{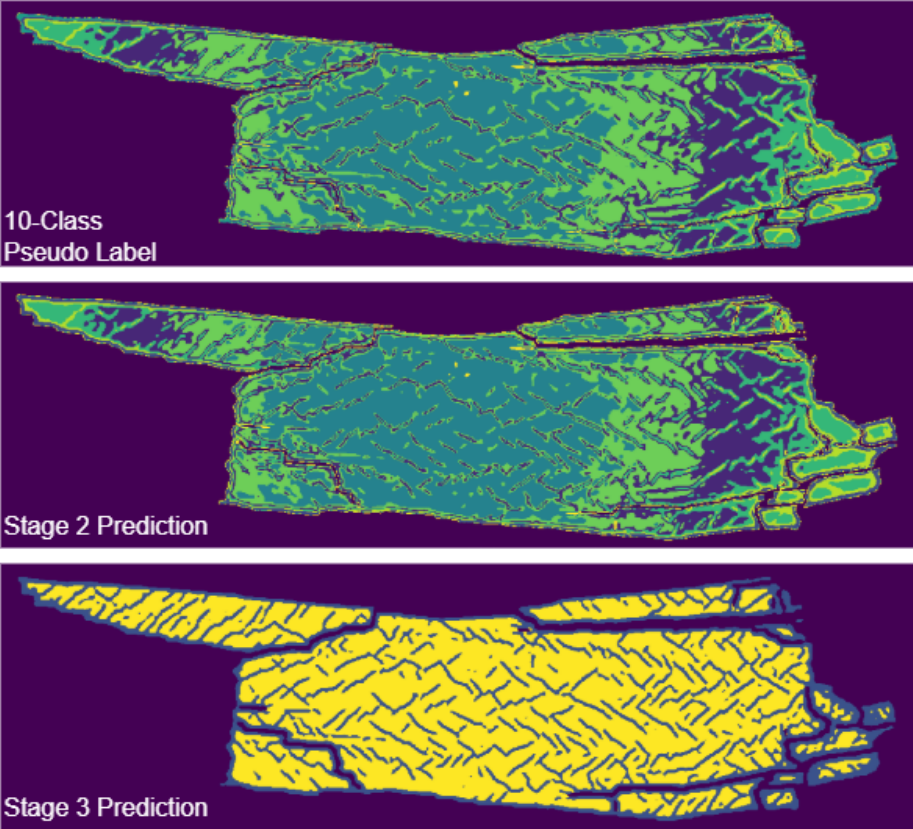}
         \caption{Results when using overly segmented pseudo label consisting of 10 classes including pseudo label, and stage 2 and 3 predictions. 
         }\label{fig:e7_10C_pred}
     \end{subfigure}
     \hfill
     \begin{subfigure}[b]{\linewidth}
         \centering
         \includegraphics[width=1.\linewidth]{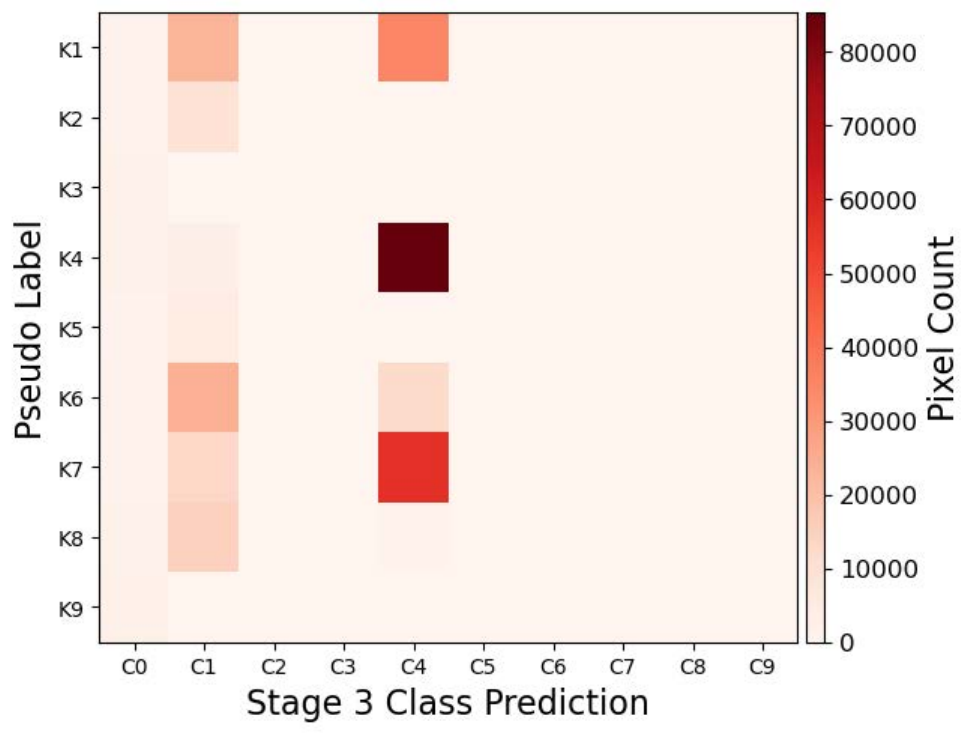}
         \caption{Confusion matrix showing the change in class assignment between the pseudo label and stage 3 when using 10 classes. Each cell represents the number of pixels that started in cluster K and ended in class C.}\label{fig:e7_10C_cm}
     \label{fig:e7_10C}
     \end{subfigure}
     \caption{Qualitative results for over-segmented SR-CT image consisting of 10 classes.}
\end{figure}

\begin{figure*}[t!]
\centering\includegraphics[width=1.0\linewidth]{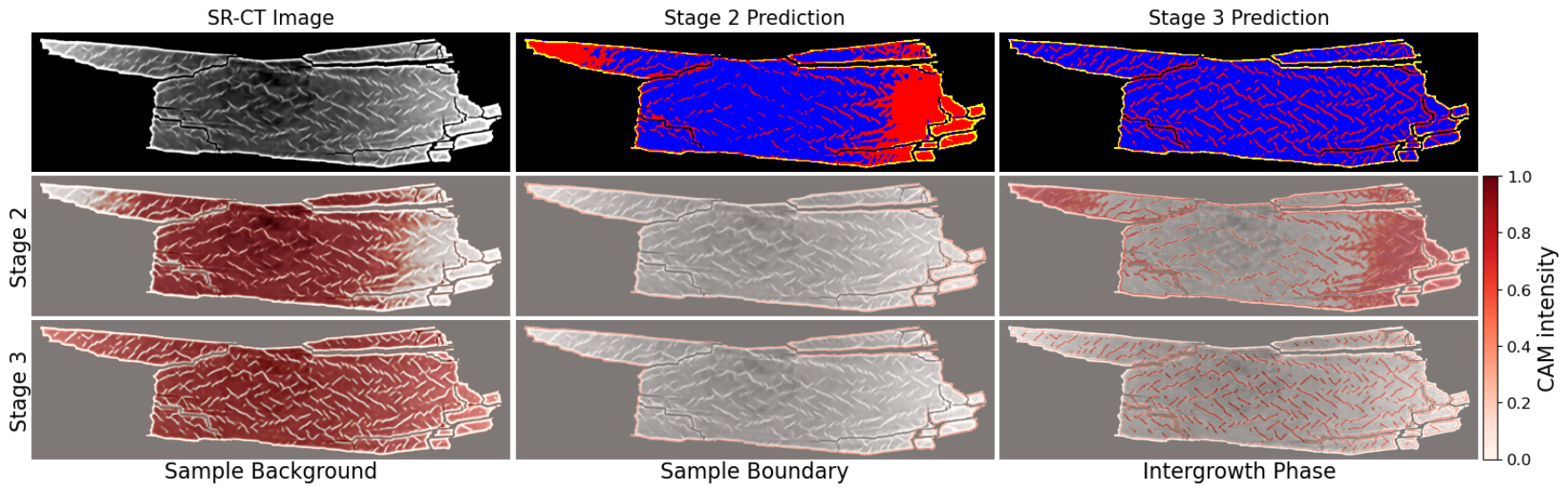}
\caption{Grad-CAM\cite{selvaraju2017grad} scores for magnesium dataset. Higher intensity values indicate regions that strongly contribute to the model's prediction for the target class.}\label{fig:ep8}
\end{figure*}

Figure \ref{fig:e7_10C_pred} shows the results when setting the number of clusters in stage 1 to 10. As expected, the pseudo label is extremely over segmented with the sample and intergrowth phases being split among numerous clusters. Again, this over segmentation presents no challenge during the initial training given the similarity between the model prediction and pseudo label. Remarkably, in stage 3, the model merges these redundant clusters into three semantically coherent classes consisting of the background, sample, and intergrowth phases. This can also be seen in Figure \ref{fig:e7_10C_cm} where the sample is initially split between clusters K1, K4, K6, and K7 before being collapsed into class C4. Similarly, the intergrowth phase is spread among clusters K1, K6, K7, K8 before being reassigned to class C1.   

\subsection{Evaluating Model Interpretability}

\begin{figure*}[t!]
\centering\includegraphics[width=1.0\linewidth]{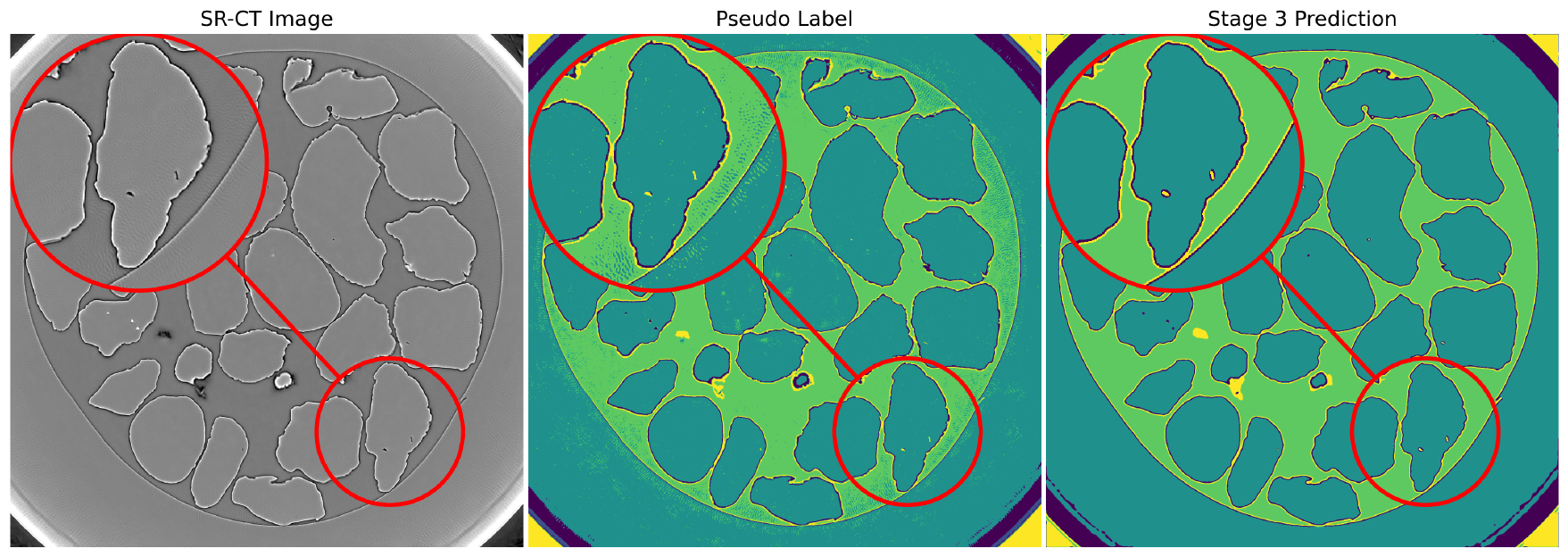}
\caption{Results of the silica sand dataset showing an example SR-CT image, pseudo label, and stage 3 prediction.  \label{fig:sand_preds}}
\end{figure*}

In addition to evaluating the different components in our framework, we also examine model interpretability to understand what changes cause the difference in model predictions between stage 2 and 3. To do this, we make use of class activation maps that highlight which regions of an input image contributed to a class score. Specifically, Grad-CAM \cite{selvaraju2017grad} weights features maps using the gradients of the predicted class's score and keeps those with a positive contribution to the class. These features are then projected and overlayed on the original image creating a heatmap enabling visualization of regions contributing to the class score. Figure \ref{fig:ep8} shows this approach applied to the magnesium sample. In the top row, we include the original SR-CT image, and stage 2 and 3 predictions. The second row includes the calculated Grad-CAM scores for the model after stage 2 with the third row showing these scores for the final model based on the sample, boundary, and intergrowth phase classes. From the figure, we observe that after stage 2, the model weights highly those regions that align with the pseudo label. However, the model assigns a slightly lower score to the intergrowth phases found inside the region on the right side of the sample, indicating that although the model assigns that region to the wrong class, it has some inclination that it may not truly belong to that class. When examining the scores after stage 3, we can see that they align much better with the ground truth structures within the sample demonstrating that the model is able to enhance its understanding of the data by identifying structures using more than just their contrast initially learned in stage 2.

\subsection{Evaluating Generalizability to Additional SR-CT Samples}

In this section, we evaluate the generalizability of our framework to the sand and ceramic SR-CT samples. Figure \ref{fig:sand_preds} depicts the results for the silica sand dataset. As mentioned above, this data was used to investigate the sand's quartz crystal structure including regions of interest such as defects, voids, and cracks. From the figure, we can see the sand grains inside the apparatus with an individual sand grain zoomed in containing two voids. The corresponding pseudo label is capable of separating the sand grains from the regions of interest and the air gaps (areas outside the grains) but suffers from noise inside the grains and a noisy texture near the apparatus. Using our framework, the final model prediction in stage 3 removes this noise and identifies a thin boundary around the two voids similar to the double boundary around the grains stemming from the fringe artifacts in the original SR-CT image.

\begin{figure*}[t!]
\centering\includegraphics[width=1.0\linewidth]{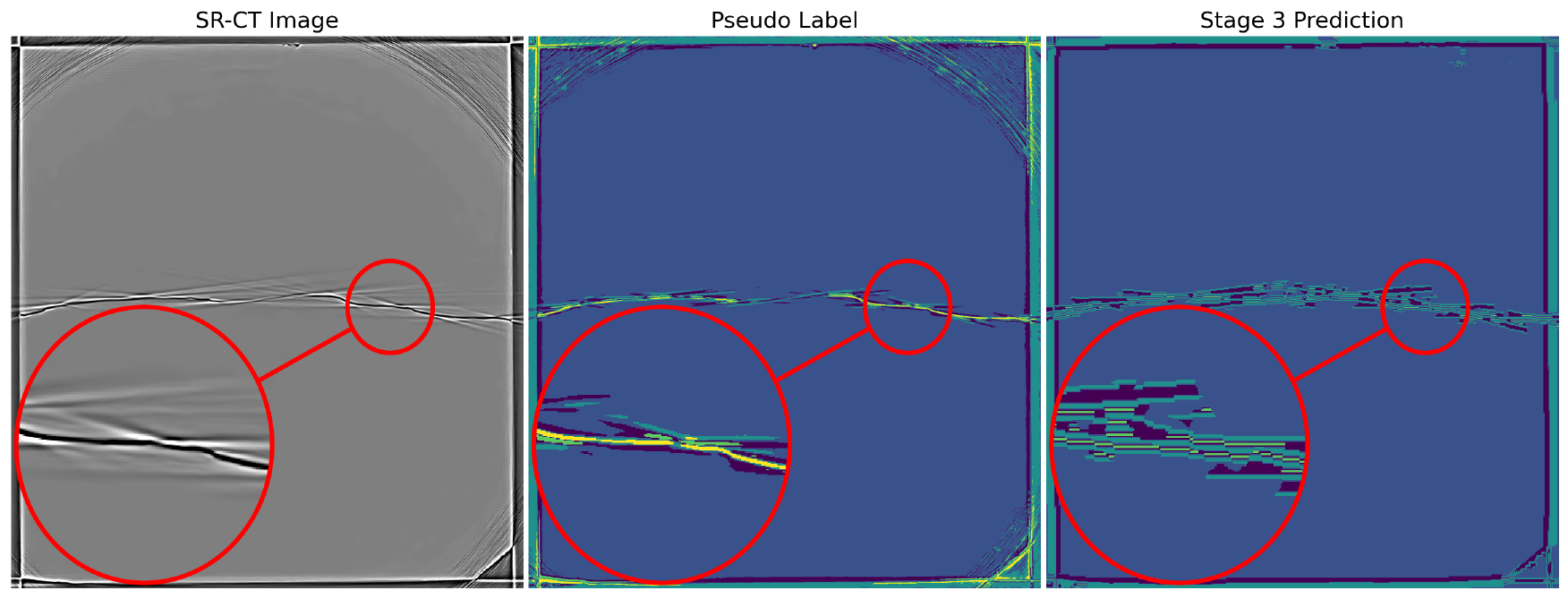}
\caption{Results of the ceramic dataset showing an SR-CT image, pseudo label, and stage 3 prediction.  \label{fig:ceramic}}
\end{figure*}

Figure \ref{fig:ceramic} shows the results when using the ceramic sample. Given the original investigation was into a pre-initiated crack within a solid AlON prismatic sample, we can see in the example SR-CT image that a majority of the sample is homogeneous with little detail. The crack can be seen running horizontally in the middle of the material with streak artifacts running tangentially from it. After clustering, the pseudo label provides an almost ideal segmentation with the ceramic, crack, and artifacts separated. However, we note that as the artifacts move further from the crack, they become less apparent and are blended into the ceramic class. After stage 3, the model prediction shows that although it can do a better job identifying the artifacts, the crack itself becomes merged with one of the artifact classes. We note that this limitation likely reflects the strong class imbalance within this sample as the ceramic class dominates both the crack and artifact classes.

\section{Discussion}
\label{sec:discussion}

In this work, we introduced a framework that enables unsupervised semantic segmentation in SR-CT samples eliminating the need to hand label images for deep learning training. Our framework contains three stages that generates pseudo labels, learns from these initial labels, and then self-corrects them. After extensively evaluating our framework and its associated components, we have come across a few noteworthy findings that we remark upon here. First, we showed that a simple U-Net with no skip connections, essentially an autoencoder, performs best in our framework which appears contradictory to recent works that demonstrate improved performance when using advanced architectures. We attribute this finding to the use of strong augmentations in stage 3. By removing the skip connections, the model is unable to pass high-level information directly to the decoder, requiring the model to learn generalizable features when faced with strong augmentations effectively boosting final segmentation accuracy. Second, although we expect the pseudo labels to contain some noise, using the standard cross entropy is sufficient to initialize the model in stage 2. The use of noise robust loss functions may provide some benefits in this stage but ultimately hurt model performance in stage 3. Furthermore, the results suggest that confidence calibration techniques offer improvements in both stages. Third, our framework requires specifying the number of clusters in the first stage to generate the pseudo labels for initial training. While methods have been developed to identify the optimal number, we saw that our approach can be robust to this choice in cases when the chosen number is close to optimal and, surprisingly, when this choice is severely over-estimated. We suggest that the number should always be set higher than expected as the model can collapse redundant classes if necessary. Fourth, by examining class activation maps, we observed that the model transitions from learning mostly contrast-related features in stage 2 to more generalizable features in stage 3 resulting in a more holistic understanding of the data. Finally, when generalizing our framework to other real-world SR-CT samples, we saw similar improvements in the final segmentations as we did with the magnesium sample. However, we recognize that our framework struggles when applied to samples with extreme class imbalance (e.g., the crack in the ceramic sample)  but we acknowledge this as a challenge inherit to all segmentation workflows. 

\section{Conclusion}
\label{sec:conc}

In this paper, we addressed the issue of segmenting large, high-resolution SR-CT datasets. We proposed a framework that allows for unsupervised semantic segmentation of these datasets by making use of pseudo labels derived from clustering the voxel values of the datasets. Although these pseudo labels provide a relevant starting point for training a deep learning model, they are commonly corrupted by noise and imaging artifacts. By adapting the Unbiased Teacher \cite{liu2021unbiased} approach for unsupervised semantic segmentation, we were able to self-correct these pseudo labels producing segmentations that better align with the ground truth labels. Overall, our framework is lightweight, avoids supervision, and adapts well to large-scale, unlabeled SR-CT datasets with the results indicating a significant enhancement in the initial pseudo labels. Our approach marks a major step towards developing an automatic segmentation workflow for SR-CT samples with minimal human involvement. Future work seeks to continuously extend this framework to additional SR-CT samples while also investigating how this approach can be adapted into a Vision Foundation Model enabling the framework to perform various vision tasks relevant to SR-CT analysis.

\section*{Acknowledgements}
This research used resources of the Advanced Photon Source, a U.S.\ Department of Energy (DOE) Office of Science user facility at Argonne National Laboratory and is based on research supported by the DOE Office of Science-Basic Energy Sciences, under Contract No. DE-AC02-06CH11357. This work was also supported by the DOE Office of Science, Office of Basic Energy Sciences Data, Artificial Intelligence and Machine Learning at DOE Scientific User Facilities program under Award Number 08735 (``Actionable Information from Sensor to Data Center (AISDC)"). The authors acknowledge Prof. Khalid Alshibli's group's (University of Tennessee), Jayden C. Plumb's (University of California) and  Zahir Islam's (Advance Photon Source) contribution by sharing their tomography data sets for demonstration of the performance of this image processing study with great experimental examples.

\section*{License}

The submitted manuscript has been created by UChicago Argonne, LLC, Operator of Argonne National Laboratory (``Argonne''). Argonne, a U.S. Department of Energy Office of Science laboratory, is operated under Contract No. DE-AC02-06CH11357. The U.S. Government retains for itself, and others acting on its behalf, a paid-up nonexclusive, irrevocable worldwide license in said article to reproduce, prepare derivative works, distribute copies to the public, and perform publicly and display publicly, by or on behalf of the Government.  The Department of Energy will provide public access to these results of federally sponsored research in accordance with the DOE Public Access Plan. http://energy.gov/downloads/doe-public-access-plan.

\vspace{12pt}

\bibliographystyle{ieeetr}  
\bibliography{main}

\begin{thebibliography}{10}

\bibitem{buzug2008computed}
T.~M. Buzug, ``Computed tomography from photon statistics to modern cone-beam ct,'' 2008.

\bibitem{omori2023recent}
N.~E. Omori, A.~D. Bobitan, A.~Vamvakeros, A.~M. Beale, and S.~D. Jacques, ``Recent developments in x-ray diffraction/scattering computed tomography for materials science,'' {\em Philosophical Transactions of the Royal Society A}, vol.~381, no.~2259, p.~20220350, 2023.

\bibitem{hiriyannaiah1997x}
H.~P. Hiriyannaiah, ``X-ray computed tomography for medical imaging,'' {\em IEEE signal Processing magazine}, vol.~14, no.~2, pp.~42--59, 1997.

\bibitem{taina2008application}
I.~Taina, R.~Heck, and T.~Elliot, ``Application of x-ray computed tomography to soil science: A literature review,'' {\em Canadian Journal of Soil Science}, vol.~88, no.~1, pp.~1--19, 2008.

\bibitem{nateghi2024health}
F.~Nateghi~Alahi and S.~Kamali, ``Health monitoring of structural elements using ct-xray,'' {\em Sharif Journal of Civil Engineering}, vol.~40, no.~3, pp.~84--92, 2024.

\bibitem{THOMPSON1984319}
A.~Thompson, J.~Llacer, L.~{Campbell Finman}, E.~Hughes, J.~Otis, S.~Wilson, and H.~Zeman, ``Computed tomography using synchrotron radiation,'' {\em Nuclear Instruments and Methods in Physics Research}, vol.~222, no.~1, pp.~319--323, 1984.

\bibitem{KIM198744}
K.-J. Kim, ``Brightness and coherence of synchrotron radiation and high-gain free electron lasers,'' {\em Nuclear Instruments and Methods in Physics Research Section A: Accelerators, Spectrometers, Detectors and Associated Equipment}, vol.~261, no.~1, pp.~44--53, 1987.

\bibitem{KASTNER2010599}
J.~Kastner, B.~Harrer, G.~Requena, and O.~Brunke, ``A comparative study of high resolution cone beam x-ray tomography and synchrotron tomography applied to fe- and al-alloys,'' {\em NDT `I\&' E International}, vol.~43, no.~7, pp.~599--605, 2010.

\bibitem{PEGUES2021159505}
J.~W. Pegues, M.~A. Melia, M.~A. Rodriguez, T.~F. Babuska, B.~Gould, N.~Argibay, A.~Greco, and A.~B. Kustas, ``In situ synchrotron x-ray imaging and mechanical properties characterization of additively manufactured high-entropy alloy composites,'' {\em Journal of Alloys and Compounds}, vol.~876, p.~159505, 2021.

\bibitem{tavakoli2020comparison}
S.~Tavakoli~Taba, P.~Baran, Y.~I. Nesterets, S.~Pacile, S.~Wienbeck, C.~Dullin, K.~Pavlov, A.~Maksimenko, D.~Lockie, S.~C. Mayo, {\em et~al.}, ``Comparison of propagation-based ct using synchrotron radiation and conventional cone-beam ct for breast imaging,'' {\em European radiology}, vol.~30, no.~5, pp.~2740--2750, 2020.

\bibitem{patterson2020novel}
C.~Patterson, D.~Murphy, S.~Irvine, L.~Connor, and Z.~Rattray, ``Novel application of synchrotron x-ray computed tomography for ex-vivo imaging of subcutaneously injected polymeric microsphere suspension formulations,'' {\em Pharmaceutical Research}, vol.~37, no.~6, p.~97, 2020.

\bibitem{hidayetouglu2020petascale}
M.~Hidayeto{\u{g}}lu, T.~Bicer, S.~G. De~Gonzalo, B.~Ren, V.~De~Andrade, D.~Gursoy, R.~Kettimuthu, I.~T. Foster, and W.-m.~W. Hwu, ``Petascale xct: 3d image reconstruction with hierarchical communications on multi-gpu nodes,'' in {\em SC20: International Conference for High Performance Computing, Networking, Storage and Analysis}, pp.~1--13, IEEE, 2020.

\bibitem{hampai2025x}
D.~Hampai, F.~Galdenzi, E.~Capitolo, A.~Di~Filippo, and S.~Dabagov, ``X-ray computed tomography—from synchrotron to desktop applications,'' {\em Nuclear Instruments and Methods in Physics Research Section A: Accelerators, Spectrometers, Detectors and Associated Equipment}, p.~170544, 2025.

\bibitem{yakovlev2022wide}
M.~A. Yakovlev, D.~J. Vanselow, M.~S. Ngu, C.~R. Zaino, S.~R. Katz, Y.~Ding, D.~Parkinson, S.~Y. Wang, K.~C. Ang, P.~La~Riviere, {\em et~al.}, ``A wide-field micro-computed tomography detector: micron resolution at half-centimetre scale,'' {\em Synchrotron Radiation}, vol.~29, no.~2, pp.~505--514, 2022.

\bibitem{khounsary2013high}
A.~Khounsary, P.~Kenesei, J.~Collins, G.~Navrotski, and J.~Nudell, ``High energy x-ray micro-tomography for the characterization of thermally fatigued glidcop specimen,'' in {\em Journal of Physics: Conference Series}, vol.~425, p.~212015, IOP Publishing, 2013.

\bibitem{han2017volume}
D.~Han, M.~A. Heuvelmans, and M.~Oudkerk, ``Volume versus diameter assessment of small pulmonary nodules in ct lung cancer screening,'' {\em Translational lung cancer research}, vol.~6, no.~1, p.~52, 2017.

\bibitem{ziegler2021computed}
F.~R. Ziegler-Rivera, B.~Prado, A.~Gastelum-Strozzi, J.~M{\'a}rquez, L.~Mora, A.~Robles, and B.~Gonz{\'a}lez, ``Computed tomography assessment of soil and sediment porosity modifications from exposure to an acid copper sulfate solution,'' {\em Journal of South American Earth Sciences}, vol.~108, p.~103194, 2021.

\bibitem{choi2024development}
W.~Choi, C.-H. Kim, H.~Yoo, H.~R. Yun, D.-W. Kim, and J.~W. Kim, ``Development and validation of a reliable method for automated measurements of psoas muscle volume in ct scans using deep learning-based segmentation: a cross-sectional study,'' {\em BMJ open}, vol.~14, no.~5, p.~e079417, 2024.

\bibitem{munir2019cancer}
K.~Munir, H.~Elahi, A.~Ayub, F.~Frezza, and A.~Rizzi, ``Cancer diagnosis using deep learning: a bibliographic review,'' {\em Cancers}, vol.~11, no.~9, p.~1235, 2019.

\bibitem{mao2023cross}
A.~Mao, M.~Mohri, and Y.~Zhong, ``Cross-entropy loss functions: Theoretical analysis and applications,'' in {\em International conference on Machine learning}, pp.~23803--23828, PMLR, 2023.

\bibitem{isensee2021nnu}
F.~Isensee, P.~F. Jaeger, S.~A. Kohl, J.~Petersen, and K.~H. Maier-Hein, ``nnu-net: a self-configuring method for deep learning-based biomedical image segmentation,'' {\em Nature methods}, vol.~18, no.~2, pp.~203--211, 2021.

\bibitem{ma2024segment}
J.~Ma, Y.~He, F.~Li, L.~Han, C.~You, and B.~Wang, ``Segment anything in medical images,'' {\em Nature Communications}, vol.~15, no.~1, p.~654, 2024.

\bibitem{nikitin2020dynamic}
V.~V. Nikitin, G.~A. Dugarov, A.~A. Duchkov, M.~I. Fokin, A.~N. Drobchik, P.~D. Shevchenko, F.~De~Carlo, and R.~Mokso, ``Dynamic in-situ imaging of methane hydrate formation and self-preservation in porous media,'' {\em Marine and Petroleum Geology}, vol.~115, p.~104234, 2020.

\bibitem{van2020survey}
J.~E. Van~Engelen and H.~H. Hoos, ``A survey on semi-supervised learning,'' {\em Machine learning}, vol.~109, no.~2, pp.~373--440, 2020.

\bibitem{chen2021semi}
X.~Chen, Y.~Yuan, G.~Zeng, and J.~Wang, ``Semi-supervised semantic segmentation with cross pseudo supervision,'' in {\em Proceedings of the IEEE/CVF conference on computer vision and pattern recognition}, pp.~2613--2622, 2021.

\bibitem{wang2022semi}
Y.~Wang, H.~Wang, Y.~Shen, J.~Fei, W.~Li, G.~Jin, L.~Wu, R.~Zhao, and X.~Le, ``Semi-supervised semantic segmentation using unreliable pseudo-labels,'' in {\em Proceedings of the IEEE/CVF conference on computer vision and pattern recognition}, pp.~4248--4257, 2022.

\bibitem{lee2013pseudo}
D.-H. Lee {\em et~al.}, ``Pseudo-label: The simple and efficient semi-supervised learning method for deep neural networks,'' in {\em Workshop on challenges in representation learning, ICML}, vol.~3, p.~896, Atlanta, 2013.

\bibitem{xu2022semi}
H.~Xu, L.~Liu, Q.~Bian, and Z.~Yang, ``Semi-supervised semantic segmentation with prototype-based consistency regularization,'' {\em Advances in neural information processing systems}, vol.~35, pp.~26007--26020, 2022.

\bibitem{yang2022st++}
L.~Yang, W.~Zhuo, L.~Qi, Y.~Shi, and Y.~Gao, ``St++: Make self-training work better for semi-supervised semantic segmentation,'' in {\em Proceedings of the IEEE/CVF conference on computer vision and pattern recognition}, pp.~4268--4277, 2022.

\bibitem{arazo2020pseudo}
E.~Arazo, D.~Ortego, P.~Albert, N.~E. O’Connor, and K.~McGuinness, ``Pseudo-labeling and confirmation bias in deep semi-supervised learning,'' in {\em 2020 International joint conference on neural networks (IJCNN)}, pp.~1--8, IEEE, 2020.

\bibitem{wang2022generalizing}
J.~Wang {\em et~al.}, ``Generalizing to unseen domains: A survey on domain generalization,'' {\em IEEE transactions on knowledge and data engineering}, vol.~35, no.~8, pp.~8052--8072, 2022.

\bibitem{liu2021unbiased}
Y.-C. Liu, C.-Y. Ma, Z.~He, C.-W. Kuo, K.~Chen, P.~Zhang, B.~Wu, Z.~Kira, and P.~Vajda, ``Unbiased teacher for semi-supervised object detection,'' {\em arXiv preprint arXiv:2102.09480}, 2021.

\bibitem{selvaraju2017grad}
R.~R. Selvaraju, M.~Cogswell, A.~Das, R.~Vedantam, D.~Parikh, and D.~Batra, ``Grad-cam: Visual explanations from deep networks via gradient-based localization,'' in {\em Proceedings of the IEEE international conference on computer vision}, pp.~618--626, 2017.

\bibitem{cooley2021semantic}
V.~Cooley {\em et~al.}, ``Semantic segmentation of mouse jaws using convolutional neural networks,'' in {\em Developments in X-Ray Tomography XIII}, vol.~11840, p.~28, SPIE, 2021.

\bibitem{badran2020automated}
A.~Badran {\em et~al.}, ``Automated segmentation of computed tomography images of fiber-reinforced composites by deep learning,'' {\em Journal of Materials Science}, vol.~55, pp.~16273--16289, 2020.

\bibitem{tsamos2023synthetic}
A.~Tsamos, S.~Evsevleev, R.~Fioresi, F.~Faglioni, and G.~Bruno, ``Synthetic data generation for automatic segmentation of x-ray computed tomography reconstructions of complex microstructures,'' {\em Journal of Imaging}, vol.~9, no.~2, p.~22, 2023.

\bibitem{manchester2025leveraging}
T.~Manchester, A.~Anders, J.~Spadotto, H.~Eccleston, W.~Beavan, H.~Arcis, and B.~J. Connolly, ``Leveraging modified ex situ tomography data for segmentation of in situ synchrotron x-ray computed tomography,'' {\em arXiv preprint arXiv:2504.19200}, 2025.

\bibitem{schwonberg2025domain}
M.~Schwonberg and H.~Gottschalk, ``Domain generalization for semantic segmentation: A survey,'' in {\em Proceedings of the IEEE/CVF Conference on Computer Vision and Pattern Recognition}, pp.~6437--6448, 2025.

\bibitem{tarvainen2017mean}
A.~Tarvainen and H.~Valpola, ``Mean teachers are better role models: Weight-averaged consistency targets improve semi-supervised deep learning results,'' {\em Advances in neural information processing systems}, vol.~30, 2017.

\bibitem{sohn2020fixmatch}
K.~Sohn {\em et~al.}, ``Fixmatch: Simplifying semi-supervised learning with consistency and confidence,'' {\em Advances in neural information processing systems}, vol.~33, pp.~596--608, 2020.

\bibitem{li2021residual}
H.~Li and H.~Zheng, ``A residual correction approach for semi-supervised semantic segmentation,'' in {\em Pattern Recognition and Computer Vision: 4th Chinese Conference, PRCV 2021, Beijing, China, October 29--November 1, 2021, Proceedings, Part IV 4}, pp.~90--102, Springer, 2021.

\bibitem{yuan2021simple}
J.~Yuan, Y.~Liu, C.~Shen, Z.~Wang, and H.~Li, ``A simple baseline for semi-supervised semantic segmentation with strong data augmentation,'' in {\em Proceedings of the IEEE/CVF International Conference on Computer Vision}, pp.~8229--8238, 2021.

\bibitem{chen2021complexmix}
Y.~Chen, X.~Ouyang, K.~Zhu, and G.~Agam, ``Complexmix: Semi-supervised semantic segmentation via mask-based data augmentation,'' in {\em 2021 IEEE International Conference on Image Processing (ICIP)}, pp.~2264--2268, IEEE, 2021.

\bibitem{wang2022learning}
Y.~Wang, J.~Zhang, M.~Kan, and S.~Shan, ``Learning pseudo labels for semi-and-weakly supervised semantic segmentation,'' {\em Pattern Recognition}, vol.~132, p.~108925, 2022.

\bibitem{kong2023pruning}
H.~Kong, G.-H. Lee, S.~Kim, and S.-W. Lee, ``Pruning-guided curriculum learning for semi-supervised semantic segmentation,'' in {\em Proceedings of the IEEE/CVF Winter Conference on Applications of Computer Vision}, pp.~5914--5923, 2023.

\bibitem{awais2025foundation}
M.~Awais, M.~Naseer, S.~Khan, R.~M. Anwer, H.~Cholakkal, M.~Shah, M.-H. Yang, and F.~S. Khan, ``Foundation models defining a new era in vision: a survey and outlook,'' {\em IEEE Transactions on Pattern Analysis and Machine Intelligence}, 2025.

\bibitem{dosovitskiy2020image}
A.~Dosovitskiy, ``An image is worth 16x16 words: Transformers for image recognition at scale,'' {\em arXiv preprint arXiv:2010.11929}, 2020.

\bibitem{zhang2023faster}
C.~Zhang, D.~Han, Y.~Qiao, J.~U. Kim, S.-H. Bae, S.~Lee, and C.~S. Hong, ``Faster segment anything: Towards lightweight sam for mobile applications,'' {\em arXiv preprint arXiv:2306.14289}, 2023.

\bibitem{ke2023segment}
L.~Ke, M.~Ye, M.~Danelljan, Y.-W. Tai, C.-K. Tang, F.~Yu, {\em et~al.}, ``Segment anything in high quality,'' {\em Advances in Neural Information Processing Systems}, vol.~36, pp.~29914--29934, 2023.

\bibitem{shaharabany2023autosam}
T.~Shaharabany, A.~Dahan, R.~Giryes, and L.~Wolf, ``Autosam: Adapting sam to medical images by overloading the prompt encoder,'' {\em arXiv preprint arXiv:2306.06370}, 2023.

\bibitem{yang2023track}
J.~Yang, M.~Gao, Z.~Li, S.~Gao, F.~Wang, and F.~Zheng, ``Track anything: Segment anything meets videos,'' {\em arXiv preprint arXiv:2304.11968}, 2023.

\bibitem{wang2023caption}
T.~Wang, J.~Zhang, J.~Fei, H.~Zheng, Y.~Tang, Z.~Li, M.~Gao, and S.~Zhao, ``Caption anything: Interactive image description with diverse multimodal controls,'' {\em arXiv preprint arXiv:2305.02677}, 2023.

\bibitem{sahoo2024unveiling}
P.~Sahoo, P.~Meharia, A.~Ghosh, S.~Saha, V.~Jain, and A.~Chadha, ``Unveiling hallucination in text, image, video, and audio foundation models: A comprehensive review,'' 2024.

\bibitem{jiao2024learning}
R.~Jiao, Y.~Zhang, L.~Ding, B.~Xue, J.~Zhang, R.~Cai, and C.~Jin, ``Learning with limited annotations: a survey on deep semi-supervised learning for medical image segmentation,'' {\em Computers in Biology and Medicine}, vol.~169, p.~107840, 2024.

\bibitem{bezak2021johns}
E.~Bezak, A.~H. Beddoe, L.~G. Marcu, M.~Ebert, and R.~Price, {\em Johns and Cunningham's the Physics of Radiology}.
\newblock Charles C Thomas Publisher, 2021.

\bibitem{Jin2010}
X.~Jin and J.~Han, {\em K-Means Clustering}, pp.~563--564.
\newblock Boston, MA: Springer US, 2010.

\bibitem{plumb2023dark}
J.~Plumb {\em et~al.}, ``Dark field x-ray microscopy below liquid-helium temperature: The case of namno2,'' {\em Materials Characterization}, vol.~204, p.~113174, 2023.

\bibitem{imseeh2020influence}
W.~H. Imseeh, K.~A. Alshibli, A.~Moslehy, P.~Kenesei, and H.~Sharma, ``Influence of crystal structure on constitutive anisotropy of silica sand at particle-scale,'' {\em Computers and Geotechnics}, vol.~126, p.~103718, 2020.

\bibitem{wu2018group}
Y.~Wu and K.~He, ``Group normalization,'' in {\em Proceedings of the European conference on computer vision (ECCV)}, pp.~3--19, 2018.

\bibitem{paszke2019pytorch}
A.~Paszke, S.~Gross, F.~Massa, A.~Lerer, J.~Bradbury, G.~Chanan, T.~Killeen, Z.~Lin, N.~Gimelshein, L.~Antiga, {\em et~al.}, ``Pytorch: An imperative style, high-performance deep learning library,'' {\em Advances in neural information processing systems}, vol.~32, 2019.

\bibitem{kingma2014}
D.~P. Kingma and J.~Ba, ``Adam: A method for stochastic optimization.'' {\itshape ArXiv e-prints}, 2014.

\bibitem{info11020125}
A.~Buslaev, V.~I. Iglovikov, E.~Khvedchenya, A.~Parinov, M.~Druzhinin, and A.~A. Kalinin, ``Albumentations: Fast and flexible image augmentations,'' {\em Information}, vol.~11, no.~2, 2020.

\bibitem{ronneberger2015u}
O.~Ronneberger, P.~Fischer, and T.~Brox, ``U-net: Convolutional networks for biomedical image segmentation,'' in {\em Medical image computing and computer-assisted intervention--MICCAI 2015: 18th international conference, Munich, Germany, October 5-9, 2015, proceedings, part III 18}, pp.~234--241, Springer, 2015.

\bibitem{zhou2018unet++}
Z.~Zhou, M.~M. Rahman~Siddiquee, N.~Tajbakhsh, and J.~Liang, ``Unet++: A nested u-net architecture for medical image segmentation,'' in {\em International workshop on deep learning in medical image analysis}, pp.~3--11, Springer, 2018.

\bibitem{huang2020unet}
H.~Huang, L.~Lin, R.~Tong, H.~Hu, Q.~Zhang, Y.~Iwamoto, X.~Han, Y.-W. Chen, and J.~Wu, ``Unet 3+: A full-scale connected unet for medical image segmentation,'' in {\em ICASSP 2020-2020 IEEE international conference on acoustics, speech and signal processing (ICASSP)}, pp.~1055--1059, Ieee, 2020.

\bibitem{zhang2018road}
Z.~Zhang, Q.~Liu, and Y.~Wang, ``Road extraction by deep residual u-net,'' {\em IEEE Geoscience and Remote Sensing Letters}, vol.~15, no.~5, pp.~749--753, 2018.

\bibitem{jha2019resunet++}
D.~Jha, P.~H. Smedsrud, M.~A. Riegler, D.~Johansen, T.~De~Lange, P.~Halvorsen, and H.~D. Johansen, ``Resunet++: An advanced architecture for medical image segmentation,'' in {\em 2019 IEEE international symposium on multimedia (ISM)}, pp.~225--2255, IEEE, 2019.

\bibitem{oktay2018attention}
O.~Oktay, J.~Schlemper, L.~L. Folgoc, M.~Lee, M.~Heinrich, K.~Misawa, K.~Mori, S.~McDonagh, N.~Y. Hammerla, B.~Kainz, {\em et~al.}, ``Attention u-net: Learning where to look for the pancreas,'' {\em arXiv preprint arXiv:1804.03999}, 2018.

\bibitem{chen2018encoder}
L.-C. Chen, Y.~Zhu, G.~Papandreou, F.~Schroff, and H.~Adam, ``Encoder-decoder with atrous separable convolution for semantic image segmentation,'' in {\em Proceedings of the European conference on computer vision (ECCV)}, pp.~801--818, 2018.

\bibitem{xie2021segformer}
E.~Xie, W.~Wang, Z.~Yu, A.~Anandkumar, J.~M. Alvarez, and P.~Luo, ``Segformer: Simple and efficient design for semantic segmentation with transformers,'' {\em Advances in neural information processing systems}, vol.~34, pp.~12077--12090, 2021.

\bibitem{zhang2018generalized}
Z.~Zhang and M.~Sabuncu, ``Generalized cross entropy loss for training deep neural networks with noisy labels,'' {\em Advances in neural information processing systems}, vol.~31, 2018.

\bibitem{wang2019symmetric}
Y.~Wang, X.~Ma, Z.~Chen, Y.~Luo, J.~Yi, and J.~Bailey, ``Symmetric cross entropy for robust learning with noisy labels,'' in {\em Proceedings of the IEEE/CVF international conference on computer vision}, pp.~322--330, 2019.

\bibitem{guo2017calibration}
C.~Guo, G.~Pleiss, Y.~Sun, and K.~Q. Weinberger, ``On calibration of modern neural networks,'' in {\em International conference on machine learning}, pp.~1321--1330, PMLR, 2017.

\bibitem{szegedy2016rethinking}
C.~Szegedy, V.~Vanhoucke, S.~Ioffe, J.~Shlens, and Z.~Wojna, ``Rethinking the inception architecture for computer vision,'' in {\em Proceedings of the IEEE conference on computer vision and pattern recognition}, pp.~2818--2826, 2016.

\bibitem{reed2014training}
S.~Reed, H.~Lee, D.~Anguelov, C.~Szegedy, D.~Erhan, and A.~Rabinovich, ``Training deep neural networks on noisy labels with bootstrapping,'' {\em arXiv preprint arXiv:1412.6596}, 2014.

\bibitem{ma2020normalized}
X.~Ma {\em et~al.}, ``Normalized loss functions for deep learning with noisy labels,'' in {\em International conference on machine learning}, pp.~6543--6553, PMLR, 2020.

\bibitem{ross2017focal}
T.-Y. Ross and G.~Doll{\'a}r, ``Focal loss for dense object detection,'' in {\em proceedings of the IEEE conference on computer vision and pattern recognition}, pp.~2980--2988, 2017.

\bibitem{hatamizadeh2022unetr}
A.~Hatamizadeh, Y.~Tang, V.~Nath, D.~Yang, A.~Myronenko, B.~Landman, H.~R. Roth, and D.~Xu, ``Unetr: Transformers for 3d medical image segmentation,'' in {\em Proceedings of the IEEE/CVF winter conference on applications of computer vision}, pp.~574--584, 2022.

\bibitem{ziabari20182}
A.~Ziabari, D.~H. Ye, S.~Srivastava, K.~D. Sauer, J.-B. Thibault, and C.~A. Bouman, ``2.5 d deep learning for ct image reconstruction using a multi-gpu implementation,'' in {\em 2018 52nd Asilomar Conference on Signals, Systems, and Computers}, pp.~2044--2049, IEEE, 2018.

\bibitem{liao2001fast}
P.-S. Liao, T.-S. Chen, P.-C. Chung, {\em et~al.}, ``A fast algorithm for multilevel thresholding,'' {\em J. Inf. Sci. Eng.}, vol.~17, no.~5, pp.~713--727, 2001.

\bibitem{xu2005survey}
R.~Xu and D.~Wunsch, ``Survey of clustering algorithms,'' {\em IEEE Transactions on neural networks}, vol.~16, no.~3, pp.~645--678, 2005.

\bibitem{yuan2019research}
C.~Yuan and H.~Yang, ``Research on k-value selection method of k-means clustering algorithm,'' {\em J}, vol.~2, no.~2, pp.~226--235, 2019.

\end{thebibliography}
\end{document}